\documentclass[sigconf]{acmart}

\AtBeginDocument{%
  }

\setcopyright{acmlicensed}
\copyrightyear{2025}
\acmYear{2025}
\acmDOI{XXXXXXX.XXXXXXX}
\acmConference[KDD'25]{KDD workshop on Evaluation and Trustworthiness of Agentic and Generative AI Models
}{August 4, 2025}{Toronto, ON, Canada}

\usepackage{multicol}
\usepackage{multirow} 
\usepackage{pifont}
\usepackage{listings}
\usepackage{graphicx,calc}

\begin{document}

\title{FECT: 
Factuality Evaluation of Interpretive AI-Generated Claims in Contact Center Conversation Transcripts}


\author{Hagyeong Shin}
\affiliation{%
  \institution{Cresta}
  \city{Sunnyvale}
  \state{California}
  \country{USA}
}
\email{hagyeong.shin@cresta.ai}
\orcid{0000-0002-8598-0786}

\author{Binoy Robin Dalal}
\affiliation{%
  \institution{Cresta}
  \city{Sunnyvale}
  \state{California}
  \country{USA}
}
\email{binoy.dalal@cresta.ai}

\author{Iwona Bialynicka-Birula}
\affiliation{%
  \institution{Cresta}
  \city{Sunnyvale}
  \state{California}
  \country{USA}
}
\email{iwona.bb@cresta.ai}

\author{Navjot Matharu}
\affiliation{%
  \institution{Cresta}
  \city{Sunnyvale}
  \state{California}
  \country{USA}
}
\email{navjot@cresta.ai}

\author{Ryan Muir}
\affiliation{%
  \institution{Cresta}
  \city{Sunnyvale}
  \state{California}
  \country{USA}
}
\email{ryan.muir@cresta.ai}

\author{Xingwei Yang}
\affiliation{%
  \institution{Cresta}
  \city{Sunnyvale}
  \state{California}
  \country{USA}
}
\email{xingwei.yang@cresta.ai}

\author{Samuel W.~K.~Wong}
\affiliation{%
  \institution{Cresta, University of Waterloo}
  \city{Waterloo}
  \state{ON}
  \country{Canada}
}
\email{sam.wong@cresta.ai}
\email{samuel.wong@uwaterloo.ca}

\renewcommand{\shortauthors}{Shin et al.}

\begin{abstract}
Large language models (LLMs) are known to hallucinate, producing natural language outputs that are not grounded in the input, reference materials, or real-world knowledge. In enterprise applications where AI features support business decisions, such hallucinations can be particularly detrimental. LLMs that analyze and summarize contact center conversations introduce a unique set of challenges for factuality evaluation, because ground-truth labels often do not exist for analytical interpretations about sentiments captured in the conversation and root causes of the business problems. To remedy this, we first introduce a \textbf{3D}---\textbf{Decompose, Decouple, Detach}---paradigm in the human annotation guideline and the LLM-judges' prompt to ground the factuality labels in linguistically-informed evaluation criteria.
We then introduce \textbf{FECT}, a novel benchmark dataset for \textbf{F}actuality \textbf{E}valuation of Interpretive AI-Generated \textbf{C}laims in Contact Center Conversation \textbf{T}ranscripts, labeled under our 3D paradigm. 
Lastly, we report our findings from aligning LLM-judges on the 3D  paradigm.  
Overall, our findings contribute a new approach for automatically evaluating the factuality of outputs generated by an AI system for analyzing contact center conversations. 

\end{abstract}



\received{13 June 2025}

\maketitle

\section{Introduction}

\begin{figure*}
    \centering
    \includegraphics[width=\linewidth]{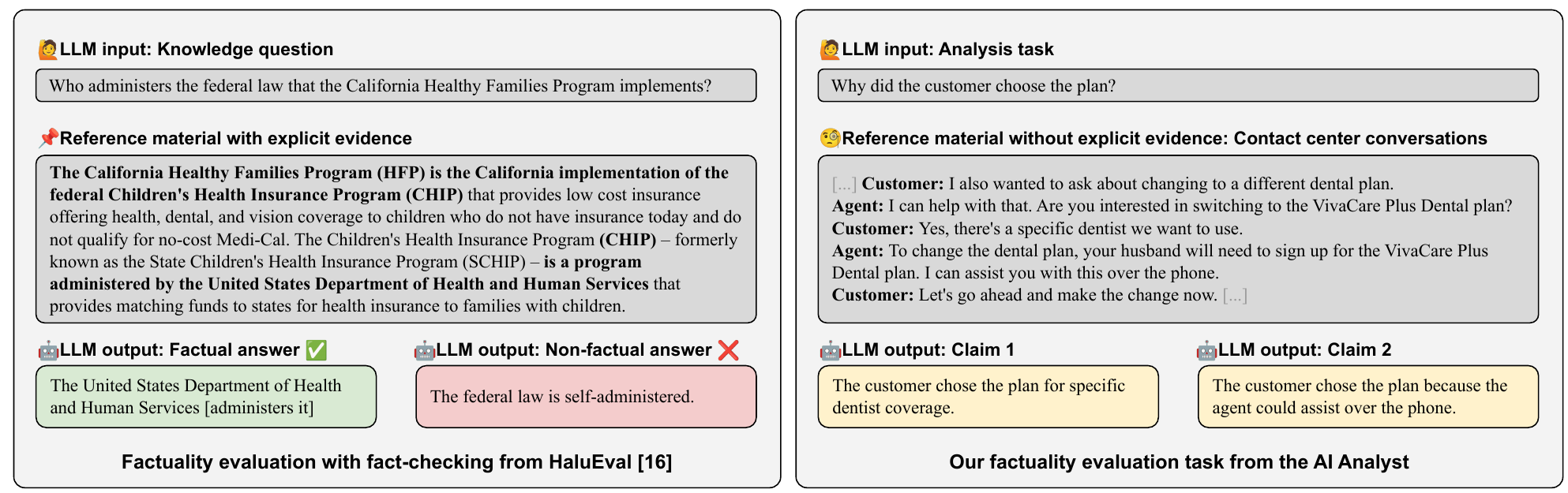}
    \caption{Most factuality evaluation tasks require fact-checking evidence, as shown on the left \cite{li_halueval_2023}. In contrast, our evaluation task on the right requires comprehensive consideration of the conversation context which does not present explicit evidence. Claim 2 is not found in our benchmark, because it is presented solely for the purpose of illustrating a non-factual claim to contrast with Claim 1. For the benchmark, we generated only one claim per conversation which may or may not be considered factual by human evaluators. See Figure \ref{fig:hallucination} for an example of a non-factual claim found in our benchmark dataset.}
    \label{fig:halueval-ours}
\end{figure*}

It remains a significant challenge to evaluate the truthfulness of outputs generated by large language models (LLMs) and LLM-based AI agents \cite{liu_g-eval_2023, wei_browsecomp_2025}. 
The most direct way of detecting LLMs' hallucinations is to have humans manually evaluate (``judge'') every LLM-generated response; however, this is a very labor intensive process, which prohibits scaling to large datasets, or rapidly iterating on the AI system's quality. A direction that has been explored is the development of automated systems that leverage LLMs themselves---an approach commonly referred to as ``LLM-as-a-Judge'' or ``LLM-Judge'' \cite{bavaresco_llms_2024, chiang_closer_2023, liu_g-eval_2023, zheng_judging_2023, kim_prometheus_2024}.
This is a promising direction to take, yet we observed challenges with our domain-specific evaluation tasks, namely that some evaluation materials are inherently ambiguous in the factuality dimension, making it challenging to establish ground-truth factuality labels.

Our factuality evaluation task originates from an enterprise AI feature developed to perform business analysis tasks---Cresta's \textsc{ai analyst} (see the right panel in Figure \ref{fig:halueval-ours}).\footnote{All conversations presented as examples in this paper are synthetically generated to reproduce patterns of original conversations while adhering to Cresta's data governance policy.} Developed by the authors, AI Analyst leverages LLMs to respond to enterprise users' research questions about their contact center conversations.
The input to AI Analyst is a user-provided \textsc{analysis task} that seeks insightful information from these conversations; for example, \textit{Why did the customer choose the plan?} asked about selected healthcare-enterprise conversations.
Leveraging LLMs, AI Analyst analyzes sampled conversations and generates a single analysis report as output. This report contains LLM-generated \textsc{claims}, which are single-sentence summaries of the conversations referenced for the analysis task. For instance, a claim \textit{The customer chose the plan for specific dentist coverage} can be the response to the analysis task aforementioned. In other words, enterprise users of our AI Analyst often request analysis tasks that require deep research, and our LLM-generated claims are often neither verbatim nor near-verbatim 
copies of the conversation; most claims are analytical interpretations made about the conversation. Thus, in the context of our AI Analyst, the factuality evaluation task is to confirm that the claim---an analytical interpretation of the conversation made in response to the analysis task---is grounded in the referenced conversation.\footnote{Our definition of factuality only takes into account the claim and the referenced conversation and does not include evaluating whether the claim adequately satisfies the analysis task. The latter is an orthogonal requirement outside the scope of this work.}



There has been substantial effort to detect hallucinations made by LLMs \cite{li_halueval_2023, lin_truthfulqa_2021, bao_faithbench_2024, luo_halludial_2024, tang_minicheck_2024, iqbal_openfactcheck_2024, wang_asking_2020}.
However, this body of work focuses on evaluating the truthfulness of LLM outputs by straightforward fact-checking illustrated in the left panel of Figure \ref{fig:halueval-ours}. Reference materials often contain explicit evidence for LLMs to extract information from, and an evaluator's task is to confirm whether the explicit evidence verifies the LLM output or not. In contrast, most contact center conversations do not contain explicit evidence to verify the LLM-generated claims. For instance, Claim 1 in Figure \ref{fig:halueval-ours} \textit{The customer chose the plan for specific dentist coverage} can be verified only when an evaluator considers the context of the customer's message \textit{Let's go ahead and make the change now} regarding the \textit{VivaCare Plus Dental plan} and the affirmative message \textit{Yes, there's a specific dentist we want to use}. In addition, the evaluator is required to judge whether the link between customer's messages and the conversational context verifies the relation stated in Claim 1, that is, the customer chose the plan specifically for the specific dentist coverage. 
Thus, the factuality evaluation of such claims involves evaluating the factuality of subjective analytical interpretations made about the conversation, which cannot be done by straightforward fact-checking. 
This nature of our AI Analyst therefore introduces a challenge seldom presented in existing hallucination detection tasks.

\begin{figure*}[t]
    \centering
    \includegraphics[width=0.7\linewidth]{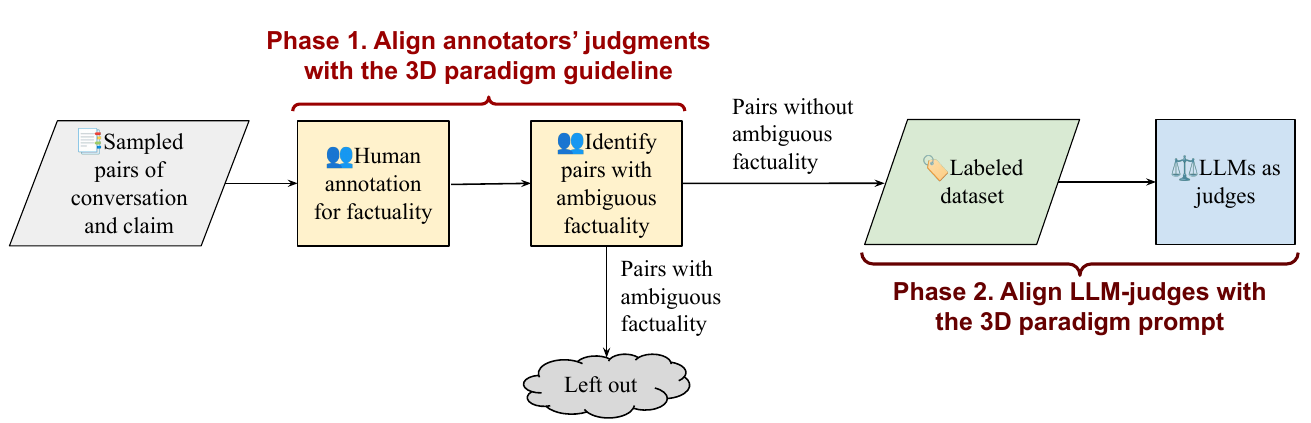}
    \caption{Overview of our methodology to develop a reliable LLM-judge for factuality evaluations. We start with Phase 1 where we align human annotators' judgments with the guideline of the 3D paradigm (described in Section \ref{sec:guideline}). We then identify tasks with ambiguous factuality to achieve a dataset with ground-truth labels. In Phase 2, we align LLM-judges with the prompt following the 3D paradigm.}
    \label{fig:method-overview}
\end{figure*}

Another challenge in our factuality evaluation task is that our LLM-generated claims about analytical interpretations of the conversation often do not yield a ground truth label of factuality.
Many evaluation tasks are inherently ambiguous \cite{aroyo_truth_2015, gordon_jury_2022, schaekermann_ambiguity-aware_2020, swayamdipta_dataset_2020, zhang_taxonomy_2023}; our tasks exhibit similar ambiguity due to the analytical nature of the claims. To our knowledge, these challenges---factuality evaluations of analytical interpretations and ambiguities in ground truth factuality---have not been addressed directly in the context of contact center conversations. 
The most related works in this type of factuality evaluation are studies that assess the factuality of dialog summarization \cite{wang_analyzing_2022, akani_increasing_2024, tang_confit_2021} and the ``extreme summarization'' dataset of news articles \cite{narayan_dont_2018, wang_asking_2020}. 
It is essential to address these challenges within the context of contact center interactions, which are characterized by distinctive features such as the use of industry-specific terminology, agent hand-offs, and the growing integration of human and AI agents.

To address this gap, we propose a methodology using LLM-judges to detect non-factual analytical claims, illustrated in Figure \ref{fig:method-overview}. Beginning with Phase 1, we first sampled pairs of conversation and claim and had human experts label the factuality of the claims. Unlike previous approaches in LLM-judge development, where granular evaluation steps only become a focus during the stage of LLM prompt iterations \cite{metropolitansky_towards_2025, tang_minicheck_2024}, we applied granular evaluation steps starting from the stage of human annotation (3D steps guideline; see Section \ref{sec:guideline}). After the factuality labeling, we identified claims for which human annotators can reach consensus on factuality, as well as those where agreement will lack due to the inherently subjective nature of the judgment. Conversation-claim pairs which humans could reasonably judge as both factual and non-factual were deliberately omitted from the dataset, since no single ground-truth label could be established for them. 
In Phase 2, we aligned LLM-judges with the prompt following the 3D paradigm so that LLM-judges are aligned on the granular evaluation process employed by human annotators. What sets our approach apart in developing reliable LLM-judges is the deliberate alignment with a structured and systematic evaluation paradigm.

Our paper thus makes the following key contributions:
\begin{enumerate}
    \item  We introduce the 3D---Decompose, Decouple, Detach---paradigm that grounds human evaluators' factuality annotations on linguistically-informed judgments and achieves the inter-annotator agreement score of 0.82.
    \item We publish FECT, a benchmark dataset for assessing factuality of interpretive claims about contact center conversations, publicly available on \texttt{\url{https://github.com/cresta/fect}}.
    \item We compare the performance of different LLM-judges on FECT and share the resulting findings. Aligning LLM-judges on the 3D paradigm can achieve a mean F1 of 0.86 without fine-tuning or extensive prompt optimizations. 
\end{enumerate}

\section{FECT dataset with human agreement}

\subsection{Data collection}
An initial dataset was created by sampling 17 analysis tasks submitted to the AI Analyst. These tasks were sampled based on salient needs that our enterprise customers across different industry verticals valued and requested. For each of the 17 tasks, 30 conversations were sampled to create LLM-generated claims. In other words, one claim was created as a summary of one conversation, creating 30 claims per analysis task, 510 in total. 
Our proprietary dataset will not be published, adhering to Cresta's data governance standard. 
Thus, we created a synthetic dataset of  conversations between fictitious customers and fictitious company contact centers that exhibit the same properties and challenges as those we observed in real contact center conversations. 

\subsection{Human evaluation guideline}\label{sec:guideline}

\begin{figure*}
    \centering
    \includegraphics[width=\linewidth]{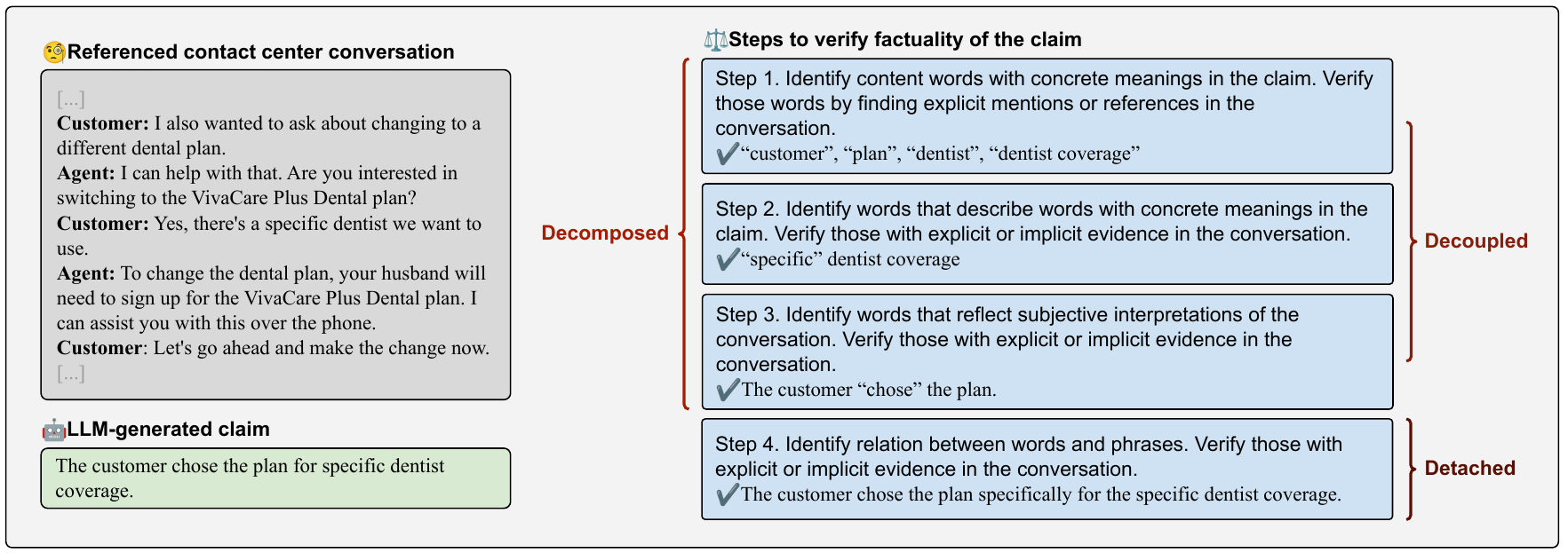}
    \caption{The visualization of our 3D guideline used by human evaluators in annotating the factuality of claims. Annotators verified information from \textbf{decomposed}  claims in Steps 1-3. Decomposed parts of claims are \textbf{decoupled} in Steps 1-3, into parts with concrete meanings and parts about subjective interpretations of the conversation. In Step 4, annotators verified the \textit{relation} between entities, which was \textbf{detached} from their meanings. In this example, information required in Steps 1 through 4 are verified by the conversation, thus the claim is judged as factual.}
    \label{fig:guideline-prompt}
\end{figure*}

From an initial round of unguided factuality labeling, we observed that different annotators had different understandings of what qualifies as ``factual'' in evaluating analytical claims. 
Thus, annotators' judgments needed to be grounded on a shared systematic evaluation paradigm rather than on each annotator's own understanding of factuality. To achieve this, we established the guideline under the \textbf{3D}---\textbf{Decompose, Decouple, Detach}---paradigm\footnote{3D paradigm is developed based on a linguistics and compositional semantics framework (see \cite{szabo_compositionality_2020} for an overview). Specific linguistics concepts utilized in the guideline include constituency, semantic decomposition, phrase structures, denotation, connotation, constitutionality, compositionality, and form-meaning correspondence.} that led annotators to ground their factuality labels on linguistically-informed evaluation criteria (Figure \ref{fig:guideline-prompt}). In the beginning of each labeling task, annotators were first instructed to \textbf{decompose} the claim into minimal informational units. In practice, annotators parsed the sentence into meaningful phrases or at word boundaries (e.g., \textit{The customer chose the plan for specific dentist coverage} $\rightarrow$ \textit{customer}, \textit{chose}, \textit{plan}, \textit{specific}, \textit{dentist coverage}, \textit{for specific dentist coverage}, etc.). 
After the decomposition of the claim, annotators \textbf{decoupled} words that have concrete meanings (e.g., \textit{plan}) from words that reflect subjective interpretations of the conversation (e.g., \textit{chose} ... (specifically) \textit{for}). 

Step 1 of the guideline (see Figure \ref{fig:guideline-prompt}) instructed to verify words of concrete meanings by finding explicit mentions or references of those words. In practice, those were often nouns and noun phrases (e.g., \textit{customer}, \textit{plan}, \textit{dentist}, \textit{dentist coverage}), which are interpreted with generally-shared meanings across English speakers and most of the times do not reflect any subjective interpretations in our claims. 
Words that reflect subjective interpretations of the conversation were verified with explicit or implicit evidence.

Step 2 of our guideline (see Figure \ref{fig:guideline-prompt})  instructed annotators to verify the words that modify the words with concrete meanings. 
In LLM-generated claims, those adjectives often reflect LLM's own interpretations of the conversation. 
Annotators were instructed to find either explicit or implicit evidence to verify these descriptive words and phrases as factual. In other words, when the conversation explicitly contains messages such as ``We want to use a specific dentist'' (explicit evidence) or  ``We want to use one particular dentist'' (\textit{specific dentist coverage} is implied), the factuality of the descriptive phrase \textit{specific} could be verified.

Step 3 in our guideline instructed to identify words about subjective interpretations made based on the entire conversation. In the context of our claims, these words were descriptions of customer's sentiment, attitudes or preferences, and behaviors that reflect those sentiments, attitudes or preferences. 
In our example,  such words were \textit{satisfaction}, \textit{confusion} or \textit{frustration} as well as  \textit{chose} in the claim \textit{The customer chose the plan for specific dentist coverage}, indicating customer's actions reflecting their preferences. Our annotators were instructed to verify these words by finding implicit evidence from the conversation. Thus, even when the customer's message does not explicitly state that ``I'll chose the plan,'' annotators found the evidence that implies customer's choosing, such as \textit{Let's go ahead and make the change now} (in Figure \ref{fig:guideline-prompt}) to verify that the claim is factual.

After verifying the decomposed and decoupled information (Steps 1-3), annotators \textbf{detached} the structure of the claim (thus relation between words) from the meaning of the claim. That is, Step 4 instructed to verify only the relation between words (who did what to whom, why and how), ignoring the exact meanings of those wh-words.
The relation was verified by finding explicit or implicit evidence. In the claim \textit{The customer chose the plan for specific dentist coverage}, the relation to verify was the causal relation between customer's choosing the plan and the specific dentist coverage. In the conversation referenced for the claim in Figure \ref{fig:guideline-prompt}, the context that the customer mentioned ``particular dentist'' and then proceeding to change the plan is considered an implied evidence showing that the relation stated in the claim is factual.

Finally, the claim was labeled as either factual or non-factual: It was labeled as factual only if all pieces of information being asked in Steps from 1 to 4 were verified. If any parts of the information could not be verified, the claim was labeled as non-factual. Applying the 3D paradigm to our guideline improved the inter-annotator agreement score between two annotators from 0.28 (Cohen's $\kappa$ observed from unguided labeling; considered as ``fair'' agreement) to 0.58 (considered as ``moderate'' agreement). Further processes to improve the agreement score is described in \ref{sec:ambiguity} and \ref{sec:fect}.

\subsection{Identifying ambiguity in human factuality evaluations}\label{sec:ambiguity}

\begin{figure*}
    \centering
    \includegraphics[width=\linewidth]{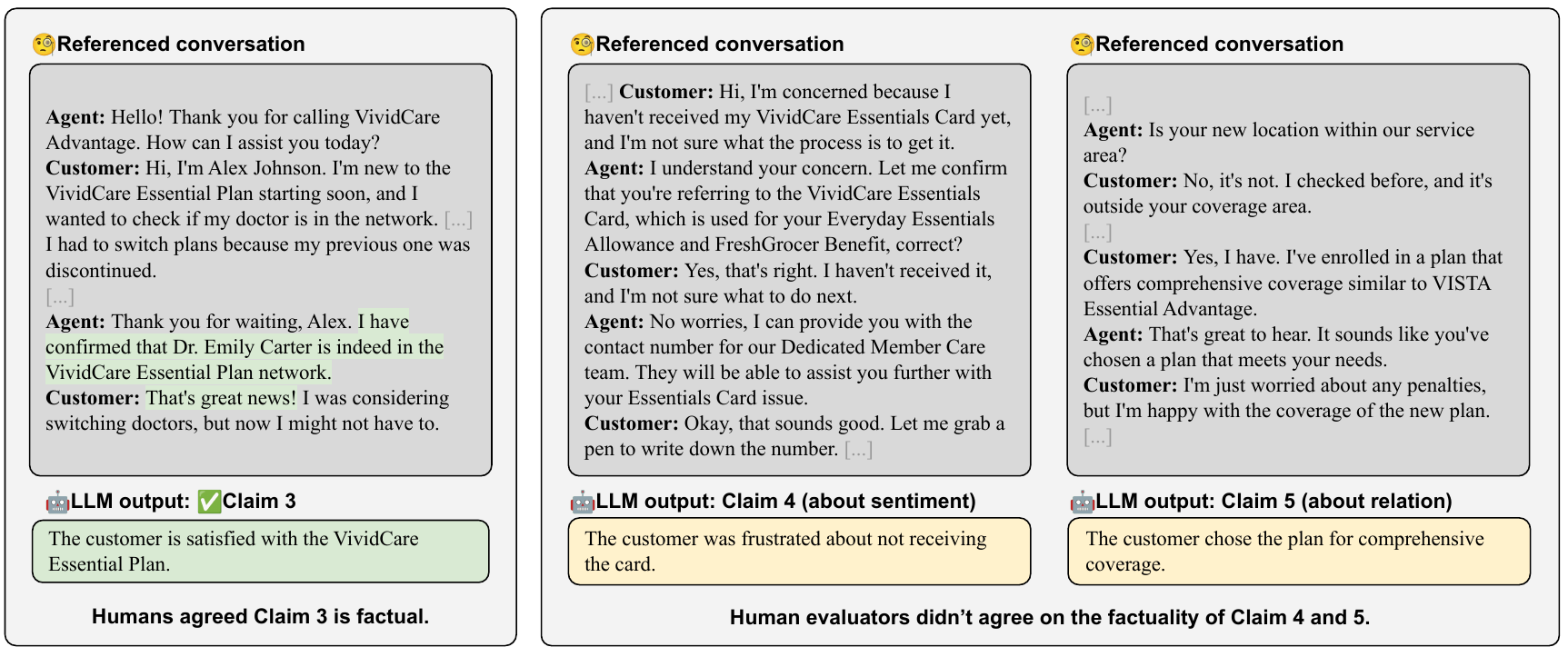}
    \caption{Claim 3 exemplifies tasks that could achieve human agreement in factuality evaluation, despite the fact that the paired conversation does not include direct references and the ``satisfaction'' is implied in highlighted messages. The conversation referenced for Claim 4 does not include messages that imply ``frustration''. The conversation referenced for Claim 5 does not include messages that imply the relation stated in Claim 5. 
    Thus, verifying Claims 4 about sentiment and Claim 5 about relation required subjective assumptions and did not yield human agreement on their factuality. }
    \label{fig:gray-area-examples}
\end{figure*}

Following the factuality labeling, annotators had multiple sessions to discuss agreement on the factuality labels. As expected, parts of claims reflecting subjective interpretations of the conversation (i.e., information needed to be verified in Steps 2-4 in our guideline and Figure \ref{fig:guideline-prompt}) were likely to introduce variances across human evaluators. As long as the evidence was implied by the actual messages in the conversation, human evaluators could reach an agreement on the claims' factuality without making subjective assumptions. For instance, Claim 3 in Figure \ref{fig:gray-area-examples} required the verification of the sentiment ``satisfaction'' and its relation with \textit{VividCare Essential Plan}. Human annotators agreed that there is a concrete message \textit{That's great news!}, uttered when they were asked about the coverage of a certain doctor under the \textit{VividCare Essential Plan}, clearly implied the customer's ``satisfaction.''

In contrast, human evaluators showed disagreement when the conversation did not contain any concrete message to ground implicit evidence so that they were led to use subjective assumptions to make judgments.
In our dataset, claims about (1) sentiment captured from the conversation and/or (2) relation between entities identified from the conversation were identified as major categories introducing ambiguity in factuality labeling. 
For instance, Claim 4 in Figure \ref{fig:gray-area-examples} doesn't have any messages where customer's frustration is implied. Some evaluators interpreted the customer's message \textit{I'm concerned because I haven't received my VividCare Essentials Card yet, and I'm not sure what the process is to get it} as an expression of a concern, not making any further assumptions. However, other evaluators assumed that the \textit{concern} could have led to the \textit{frustration} mentioned in the claim. In other words, verifying Claim 4 introduced room for evaluators to make subjective assumptions to verify ``frustration.''



Claim 5 in Figure \ref{fig:gray-area-examples} states that the root cause of the customer's choice was the \textit{comprehensive coverage}. 
Some evaluators suggested that the customer's message \textit{I'm happy with the coverage of the new plan} indicates the causal relation between customer's choice and comprehensive coverage, thus labeling the Claim 5 as factual. Other evaluators suggested that the customer's moving implied in the conversation is a more plausible cause of the customer's plan choice. In making these two interpretations, human evaluators used their own subjective assumptions to verify the relation.

\subsection{FECT benchmark dataset}\label{sec:fect}

Conversation-claim pairs that are identified as inherently ambiguous to evaluate are not desired in a benchmark dataset: Claims without a ground-truth factuality cannot indicate whether the LLM-judges' factuality labels are correct or incorrect. 
We thus excluded those ambiguous pairs from our dataset. The final agreement score of 0.82 (considered as ``almost perfect'' agreement) was achieved after we excluded ambiguous tasks. 
After we achieved the near-perfect agreement, we confirmed that our 3D guideline and ambiguity identification and reduction process (Phase 1 in Figure \ref{fig:method-overview}) indeed ensured alignment between human evaluators. The benchmark dataset of synthetic conversations was labeled by 5 human experts in the domains of ML, AI, NLP, and linguistics.
Our resulting benchmark dataset, \textbf{FECT} (\textbf{F}actuality \textbf{E}valuation of Interpretive AI-Generated \textbf{C}laims
in Contact Center Conversation \textbf{T}ranscripts), consists of 410 pairs (345 factual; 65 non-factual) of LLM-generated claims deduced from synthetically generated conversations (\url{https://github.com/cresta/fect}).\footnote{The distribution of factual and non-factual labels in the the synthetic conversation dataset is very similar to the one we observed when analyzing real enterprise customers’ use cases.
This suggests that LLM-judges developed using the synthetic dataset will likely translate into improvements in LLM-judges employed in real contact center applications.} Label distributions in our dataset are reported in Table \ref{tab:dataset-stats}.

\begin{table}[ht!]
    \centering
    \begin{tabular}{cccc} \toprule
          & Agreement achieved   & \multicolumn{2}{c}{Agreement not achieved} \\  \cmidrule{3-4}
         & \textbf{FECT} & Sentiment & Relation \\ \midrule
         Factual  & \textbf{345} &  \multirow{2}{*}{31}  &  \multirow{2}{*}{53}  \\
         Non-factual & \textbf{65} &  &  \\ \cmidrule{1-2} 
        Total & \textbf{410} & \\
        \bottomrule
    \end{tabular}
    \caption{Distributions of factual and non-factual claims in our dataset and claims that required assumption-driven judgments for factuality labeling. Our benchmark dataset consists of the boldfaced portion where factuality could be determined based on evidence-driven judgments.}
    \label{tab:dataset-stats}
\end{table}

\section{Alignment between humans and LLM-Judges}
\subsection{Experimental setup}

\begin{figure*}
    \centering
    \includegraphics[width=\linewidth]{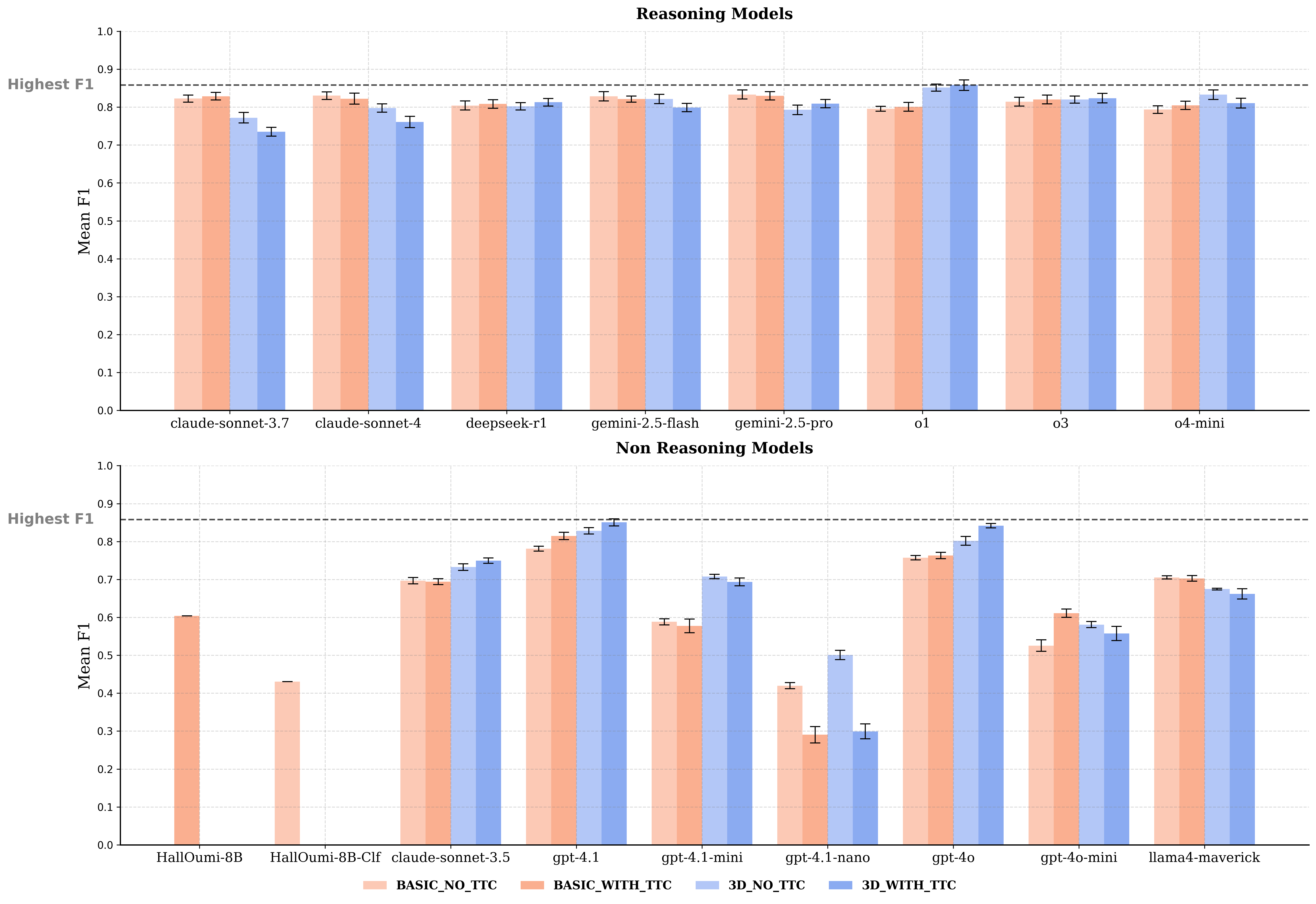}
    \caption{Means and 95\% CIs of F1s achieved with reasoning models and non-reasoning models and with four prompts. Dashed horizontal lines indicate the highest F1 of 0.86 achieved with o1 with the \texttt{3D\_WITH\_TTC} prompt. Note that our \texttt{3D} prompt was optimized for its structure and formatting only with the o1 model, and we didn't perform extensive prompt iterations with any of the models.} 
    \label{fig:f1}
\end{figure*}

The goal of our LLM-judge was to achieve optimal alignment with human evaluators not only on the final factuality labels but also on the evaluation paradigm. 
For this, we optimized the structure and the formatting of our 3D guideline for OpenAI's o1 model to obtain the \texttt{3D} prompt (see \ref{sec:3d-with-ttc} and \ref{sec:3d-no-ttc}). The prompt optimization process involved simple formatting edits without any extensive prompt iterations. 
In order to test LLM-judges' intrinsic capability to align with human evaluators' factuality labels without additional explicit reasoning cues, we compared our  \texttt{3D} prompts with \texttt{BASIC} prompts, which asked to judge the factuality without the granular 3D steps (``Given a conversation and a claim about that conversation, determine if the claim is factual, i.e., supported by the conversation'' in \ref{sec:basic-with-ttc} and \ref{sec:basic-no-ttc}).

In addition to ablating the \texttt{3D} prompt in the \texttt{BASIC} prompt, we tested prompts that allowed LLM-judges to generate intermediate reasoning tokens before generating the final factuality label, thus leveraging test-time compute (\texttt{TTC}). 
Test-time compute has been employed as a common approach to improving an LLM's accuracy, especially on tasks that require complex multi-step reasoning. The approach first emerged in the form of chain-of-thought prompting \cite{wei_chain--thought_2023} and has since become a primary mechanism for improving LLM performance \cite{zhang_survey_2025}.
Overall, our experiment tested prompts \texttt{3D\_WITH\_TTC} and \texttt{BASIC\_WITH\_TTC}, which instructed the LLM to generate intermediate outputs before the factuality label; prompts \texttt{3D\_NO\_TTC} and \texttt{BASIC\_NO\_TTC}, which did not require any intermediate outputs to be generated before the factuality label. See Appendix \ref{sec:appendix-prompts} for the 4 prompts used in our experiments. 


We tested 17 LLMs with the 4 different prompt variants described above and the results can be found in Figure \ref{fig:f1} (see Table \ref{tab:list-models} for the exact versions of the LLMs used in the experiment). We evaluated a representative mix of reasoning and non-reasoning models and 2 models fine-tuned specifically for hallucination detections (HallOumi-8B and HallOumi-8B-Classifier). Reasoning models refers to the group of models which generate internal reasoning tokens before the final output, in addition to the extra tokens explicitly elicited by the \texttt{\_WITH\_TTC} prompt variants. In order to observe the self-consistency of each model, we ran each model with each of the 4 prompts 10 times. 

\begin{table*}[h]
\centering
\begin{tabular}
{lcccc}
\toprule
 & \multicolumn{4}{c}{\textbf{F1 scores (mean ± std)}} \\
 \textbf{Model Name} &
\texttt{BASIC\_} \texttt{NO\_} \texttt{TTC} & \texttt{BASIC\_} \texttt{WITH\_} \texttt{TTC} & \texttt{3D\_} \texttt{NO\_} \texttt{TTC} & \texttt{3D\_} \texttt{WITH\_} \texttt{TTC} \\
\midrule
claude-sonnet-3.5 & 0.70 ± 0.01 & 0.69  ± 0.01 & 0.73  ± 0.01 & 0.75 ± 0.01 \\
claude-sonnet-3.7 & 0.82 ± 0.01 & \textbf{0.83} ± \textbf{0.01} & 0.77 ± 0.02 & 0.73 ± 0.02 \\
claude-sonnet-4 & \textbf{0.83}  ± \textbf{0.01} & 0.82  ± 0.01 & 0.80 ± 0.01 & 0.76 ± 0.02 \\
deepseek-r1 & 0.80 ± 0.02 & 0.81 ± 0.02 & 0.80 ± 0.01 & 0.81 ± 0.01 \\
gemini-2.5-flash & 0.83 ± 0.02 & 0.82 ± 0.01 & 0.82 ± 0.02 & 0.80 ± 0.02 \\
gemini-2.5-pro & 0.83 ± 0.02 & 0.83 ± 0.02 & 0.79 ± 0.02 & 0.81 ± 0.02 \\
gpt-4.1 & 0.78 ± 0.01 & 0.81 ± 0.01 & 0.83 ± 0.01 & 0.85 ± 0.02 \\
gpt-4.1-mini & 0.59 ± 0.01 & 0.58 ± 0.03 & 0.71 ± 0.01 & 0.69 ± 0.01 \\
gpt-4.1-nano & 0.42 ± 0.01 & 0.29 ± 0.03 & 0.50 ± 0.02 & 0.31 ± 0.04 \\
gpt-4o & 0.76 ± 0.01 & 0.76 ± 0.01 & 0.80 ± 0.02 & 0.84 ± 0.01 \\
gpt-4o-mini & 0.53 ± 0.02 & 0.61 ± 0.02 & 0.58 ± 0.01 & 0.56 ± 0.03 \\
llama4-maverick & 0.71 ± 0.01 & 0.70 ± 0.01 & 0.67 ± 0.00 & 0.67 ± 0.03 \\
o1 & 0.80 ± 0.01 & 0.80 ± 0.02 & \textbf{0.85} ± \textbf{0.01} & \textbf{0.86} \textbf{± 0.02} \\
o3 & 0.81 ± 0.02 & 0.82 ± 0.02 & 0.82 ± 0.01 & 0.82 ± 0.02 \\
o4-mini & 0.79 ± 0.02 & 0.81 ± 0.02 & 0.83 ± 0.02 & 0.81 ± 0.02 \\
HallOumi-8B & -- & 0.60 ± 0.00 & -- & -- \\
HallOumi-8B-Clf & 0.43 ± 0.00 & -- & -- & -- \\
\bottomrule
\end{tabular}
\caption{Mean and standard deviation of F1 scores across 17 models under 4 prompting modes. Boldfaced scores indicate the best mean (with the lowest standard deviation in case of a tie) within each prompt mode.}
\label{tab:f1}
\end{table*}


\subsection{Results}
F1 scores on the task of detecting non-factual claims in FECT are reported in Figure \ref{fig:f1} (see precision scores in Figure \ref{fig:precision} and recall scores in Figure \ref{fig:recall}). Numeric scores can be found in Table \ref{tab:f1}, \ref{tab:precision} and \ref{tab:recall}. 
In all tables and figures, we report the mean and 95\% confidence intervals (CIs) of scores obtained from each model per each prompt mode ($
CI_{95\%} = t_{0.025, \, df=9} \times \frac{s}{\sqrt{n}} = 2.262 \times \frac{s}{\sqrt{10}}$, $s$ = sample standard deviation of F1/Precision/Recall, $n$ = 10 runs, $df = n -1$).

Overall, reasoning models performed better with all 4 types of prompts compared to non-reasoning models (Figure \ref{fig:f1}). OpenAI's o1 model showed the best score with the \texttt{3D\_WITH\_TTC} prompt. This was expected, because our prompt iterations were performed to optimize o1's performance.  Looking into reasoning models' results first (the top row in Figure \ref{fig:f1}), 
Deepseek-r1 and o3 showed balanced scores across different prompts, while yielding slightly higher scores with \texttt{WITH\_TTC} prompts. This indicates that these models can consistently align with humans on reasoning tasks even without explicit reasoning cues provided in the prompt, while marginal improvement can be expected with the additional test-time compute step. 
Two Claude-Sonnet models and two Gemini models resulted in better scores with \texttt{BASIC} prompts. We address this result in the discussion section.




Non-reasoning models were more sensitive to the different prompting techniques, with \texttt{3D} prompts improving most models' alignment on the multi-step evaluation task. Comparing scores from \texttt{NO\_TTC} prompts, all models except Llama4-Maverick showed higher scores with the \texttt{3D} prompts than with the \texttt{BASIC} prompts. In the case of frontier models, such as Claude-Sonnet-3.5, GPT-4.1, and GPT-4o, adding test-time compute additionally boosted the models' performance, bringing the OpenAI non-reasoning models GPT-4.1 and GPT-4o almost on par with the best reasoning models. This result indicates that when explicit reasoning cues are combined with the test-time compute, non-reasoning models have the capability to align with humans' judgments, to a similar extent as reasoning models do.

Interestingly, and somewhat counterintuitively, smaller models---GPT-4.1-mini and GPT-4.1-nano (intended to approximate GPT-4.1) and GPT-4o-mini (intended to approximate GPT-4o)---showed markedly worse performance when test-time compute was added. This is especially apparent with the smallest of these models---GPT-4.1-nano---and more common with the \texttt{3D} prompt than with the \texttt{BASIC} prompt. We hypothesize that this is because the smaller models do not have enough capacity to perform the complex reasoning required by the task, so giving them the capacity to perform this reasoning not only does not improve, but can even greatly hurt performance. The key takeaway from this observation is that adding test-time compute does not automatically boost performance and can actually degrade it in the case in which the task involves complex reasoning and the model used is small.

Lastly, HallOumi-8B models which are fine-tuned specifically for hallucination detection achieved comparable scores to much larger models.
HallOumi-8B (a generative model; comparing its score with other 
\texttt{BASIC\_WITH\_TTC} scores) showed similar scores to GPT-4.1-mini and GPT-4o-mini. HallOumi-8B-Classifier (a classifier model; comparing its score with other
\texttt{BASIC\_NO\_TTC} scores) showed comparable scores to GPT-4.1-nano. This result confirms the contributions of fine-tuning reported by the developers of the models \cite{jeremy_greer_introducing_2025}.


\section{Discussion and Future Work}

We discussed that human alignment in benchmark labeling can be ensured by breaking down evaluation tasks into granular steps and by grounding judgments of each step to linguistically-informed concepts. When the ambiguity can be removed from evaluation tasks, reasoning LLMs show good alignment with the human-labeled benchmark without extensive prompt iterations or further fine-tuning. When the human evaluation process can be broken down into granular steps (as in our \texttt{3D} prompt), frontier non-reasoning models with test-time compute can reach a similar level of alignment as reasoning models.  It is our future work to optimize the alignment among humans and between humans and LLM-judges on ambiguous tasks, which were left out from our benchmark (c.f., Subsection \ref{sec:ambiguity} and Table \ref{tab:dataset-stats}). 


Among the models we tested, o1 with the \texttt{3D\_WITH\_TTC} prompt yielded the highest F1 score (0.86). Other reasoning models, such as Claude-Sonnet, Gemini, or Deepseek-r1 yielded comparable F1 scores to o1 with \texttt{BASIC} prompts (0.80--0.83), but did not benefit from the \texttt{3D} prompts as o1 did. The boost in o1's F1 scores after switching from  \texttt{BASIC} to \texttt{3D} prompts is primarily driven by striking a better balance between precision and recall. 
Since we performed prompt iterations only with the o1 model, it is possible that the other reasoning models' performances can likewise be improved by optimizing the prompts for those models. It is our future work to investigate optimizations of our \texttt{3D} prompt with different types of reasoning models other than o1.

\section{Conclusion}

As AI systems are used for tasks that require human-level intelligence, evaluating their output will also require human-level intelligence. In this paper, we presented a method to automate evaluations that involve judging the factuality of analytical interpretations about contact center conversations. Our evaluation tasks could not be done by extracting information from reference materials---instead, implicit understanding of the conversation was needed to evaluate the factuality of analysis made about the conversation. To establish the ground-truth labels for the factuality of the analytical claims, we first identified a human evaluation process that could ensure alignment between human evaluators. The second phase was to utilize this evaluation process in the LLM-judges' prompt. Based on our experimental results, we conclude that utilizing reasoning models results in good LLM-judge performances without further prompt iterations or fine-tuning, while using non-reasoning models with the explicit instructions and additional test-time compute offered comparable performances. We believe that our emphasis on achieving alignment between humans as a starting point for the development of an automatic evaluation system can contribute to the evaluation of LLMs and AI systems.

\section*{Acknowledgments}

We thank Motoki Wu, Fei Fang and Kai Zou for their helpful suggestions on our drafts. We are grateful to Robert Kugler and Jake McDermott for their guidance on adhering to Cresta's data security and privacy standards. Lastly, we extend our thanks to our reviewers for their valuable feedback.

\bibliographystyle{ACM-Reference-Format}
\bibliography{references}


\begin{thebibliography}{43}


\ifx \showCODEN    \undefined \def \showCODEN     #1{\unskip}     \fi
\ifx \showISBNx    \undefined \def \showISBNx     #1{\unskip}     \fi
\ifx \showISBNxiii \undefined \def \showISBNxiii  #1{\unskip}     \fi
\ifx \showISSN     \undefined \def \showISSN      #1{\unskip}     \fi
\ifx \showLCCN     \undefined \def \showLCCN      #1{\unskip}     \fi
\ifx \shownote     \undefined \def \shownote      #1{#1}          \fi
\ifx \showarticletitle \undefined \def \showarticletitle #1{#1}   \fi
\ifx \showURL      \undefined \def \showURL       {\relax}        \fi
\providecommand\bibfield[2]{#2}
\providecommand\bibinfo[2]{#2}
\providecommand\natexlab[1]{#1}
\providecommand\showeprint[2][]{arXiv:#2}

\bibitem[Akani et~al\mbox{.}(2024)]%
        {akani_increasing_2024}
\bibfield{author}{\bibinfo{person}{Eunice Akani}, \bibinfo{person}{Benoit Favre}, \bibinfo{person}{Frederic Bechet}, {and} \bibinfo{person}{Romain Gemignani}.} \bibinfo{year}{2024}\natexlab{}.
\newblock \bibinfo{title}{Increasing faithfulness in human-human dialog summarization with {Spoken} {Language} {Understanding} tasks}.
\newblock
\href{https://doi.org/10.48550/arXiv.2409.10070}{doi:\nolinkurl{10.48550/arXiv.2409.10070}}
\newblock
\shownote{arXiv:2409.10070}.


\bibitem[{Anthropic}(2024)]%
        {anthropic_claude_2024}
\bibfield{author}{\bibinfo{person}{{Anthropic}}.} \bibinfo{year}{2024}\natexlab{}.
\newblock \bibinfo{title}{Claude 3.5 {Sonnet} {Model} {Card} {Addendum}}.
\newblock
\urldef\tempurl%
\url{https://www-cdn.anthropic.com/fed9cc193a14b84131812372d8d5857f8f304c52/Model_Card_Claude_3_Addendum.pdf}
\showURL{%
\tempurl}


\bibitem[{Anthropic}(2025a)]%
        {anthropic_claude_2025}
\bibfield{author}{\bibinfo{person}{{Anthropic}}.} \bibinfo{year}{2025}\natexlab{a}.
\newblock \bibinfo{title}{Claude 3.7 {Sonnet} {System} {Card}}.
\newblock
\urldef\tempurl%
\url{https://assets.anthropic.com/m/785e231869ea8b3b/original/claude-3-7-sonnet-system-card.pdf}
\showURL{%
\tempurl}


\bibitem[{Anthropic}(2025b)]%
        {anthropic_system_2025}
\bibfield{author}{\bibinfo{person}{{Anthropic}}.} \bibinfo{year}{2025}\natexlab{b}.
\newblock \bibinfo{title}{System {Card}: {Claude} {Opus} 4 \& {Claude} {Sonnet} 4}.
\newblock
\urldef\tempurl%
\url{https://www-cdn.anthropic.com/6be99a52cb68eb70eb9572b4cafad13df32ed995.pdf}
\showURL{%
\tempurl}


\bibitem[Aroyo and Welty(2015)]%
        {aroyo_truth_2015}
\bibfield{author}{\bibinfo{person}{Lora Aroyo} {and} \bibinfo{person}{Chris Welty}.} \bibinfo{year}{2015}\natexlab{}.
\newblock \showarticletitle{Truth {Is} a {Lie}: {Crowd} {Truth} and the {Seven} {Myths} of {Human} {Annotation}}.
\newblock \bibinfo{journal}{\emph{AI Magazine}} \bibinfo{volume}{36}, \bibinfo{number}{1} (\bibinfo{date}{March} \bibinfo{year}{2015}), \bibinfo{pages}{15--24}.
\newblock
\showISSN{0738-4602, 2371-9621}
\href{https://doi.org/10.1609/aimag.v36i1.2564}{doi:\nolinkurl{10.1609/aimag.v36i1.2564}}


\bibitem[Bao et~al\mbox{.}(2024)]%
        {bao_faithbench_2024}
\bibfield{author}{\bibinfo{person}{Forrest~Sheng Bao}, \bibinfo{person}{Miaoran Li}, \bibinfo{person}{Renyi Qu}, \bibinfo{person}{Ge Luo}, \bibinfo{person}{Erana Wan}, \bibinfo{person}{Yujia Tang}, \bibinfo{person}{Weisi Fan}, \bibinfo{person}{Manveer~Singh Tamber}, \bibinfo{person}{Suleman Kazi}, \bibinfo{person}{Vivek Sourabh}, \bibinfo{person}{Mike Qi}, \bibinfo{person}{Ruixuan Tu}, \bibinfo{person}{Chenyu Xu}, \bibinfo{person}{Matthew Gonzales}, \bibinfo{person}{Ofer Mendelevitch}, {and} \bibinfo{person}{Amin Ahmad}.} \bibinfo{year}{2024}\natexlab{}.
\newblock \bibinfo{title}{{FaithBench}: {A} {Diverse} {Hallucination} {Benchmark} for {Summarization} by {Modern} {LLMs}}.
\newblock
\href{https://doi.org/10.48550/arXiv.2410.13210}{doi:\nolinkurl{10.48550/arXiv.2410.13210}}
\newblock
\shownote{arXiv:2410.13210}.


\bibitem[Bavaresco et~al\mbox{.}(2024)]%
        {bavaresco_llms_2024}
\bibfield{author}{\bibinfo{person}{Anna Bavaresco}, \bibinfo{person}{Raffaella Bernardi}, \bibinfo{person}{Leonardo Bertolazzi}, \bibinfo{person}{Desmond Elliott}, \bibinfo{person}{Raquel Fernández}, \bibinfo{person}{Albert Gatt}, \bibinfo{person}{Esam Ghaleb}, \bibinfo{person}{Mario Giulianelli}, \bibinfo{person}{Michael Hanna}, \bibinfo{person}{Alexander Koller}, \bibinfo{person}{André F.~T. Martins}, \bibinfo{person}{Philipp Mondorf}, \bibinfo{person}{Vera Neplenbroek}, \bibinfo{person}{Sandro Pezzelle}, \bibinfo{person}{Barbara Plank}, \bibinfo{person}{David Schlangen}, \bibinfo{person}{Alessandro Suglia}, \bibinfo{person}{Aditya~K. Surikuchi}, \bibinfo{person}{Ece Takmaz}, {and} \bibinfo{person}{Alberto Testoni}.} \bibinfo{year}{2024}\natexlab{}.
\newblock \bibinfo{title}{{LLMs} instead of {Human} {Judges}? {A} {Large} {Scale} {Empirical} {Study} across 20 {NLP} {Evaluation} {Tasks}}.
\newblock
\href{https://doi.org/10.48550/arXiv.2406.18403}{doi:\nolinkurl{10.48550/arXiv.2406.18403}}
\newblock
\shownote{arXiv:2406.18403 [cs]}.


\bibitem[Chiang and Lee(2023)]%
        {chiang_closer_2023}
\bibfield{author}{\bibinfo{person}{Cheng-Han Chiang} {and} \bibinfo{person}{Hung-yi Lee}.} \bibinfo{year}{2023}\natexlab{}.
\newblock \bibinfo{title}{A {Closer} {Look} into {Automatic} {Evaluation} {Using} {Large} {Language} {Models}}.
\newblock
\href{https://doi.org/10.48550/arXiv.2310.05657}{doi:\nolinkurl{10.48550/arXiv.2310.05657}}
\newblock
\shownote{arXiv:2310.05657 [cs]}.


\bibitem[{DeepSeek}(2025)]%
        {deepseek_deepseek-r1_2025}
\bibfield{author}{\bibinfo{person}{{DeepSeek}}.} \bibinfo{year}{2025}\natexlab{}.
\newblock \bibinfo{title}{{DeepSeek}-{R1} {Release}}.
\newblock
\urldef\tempurl%
\url{https://api-docs.deepseek.com/news/news250120}
\showURL{%
\tempurl}


\bibitem[{Google}(2025a)]%
        {google_gemini_2025}
\bibfield{author}{\bibinfo{person}{{Google}}.} \bibinfo{year}{2025}\natexlab{a}.
\newblock \bibinfo{title}{Gemini 2.5 {Flash} {Preview} {Model} {Card}}.
\newblock
\urldef\tempurl%
\url{https://storage.googleapis.com/model-cards/documents/gemini-2.5-flash-preview.pdf}
\showURL{%
\tempurl}


\bibitem[{Google}(2025b)]%
        {google_gemini_2025-1}
\bibfield{author}{\bibinfo{person}{{Google}}.} \bibinfo{year}{2025}\natexlab{b}.
\newblock \bibinfo{title}{Gemini 2.5 {Pro} {Preview} {Model} {Card}}.
\newblock
\urldef\tempurl%
\url{https://storage.googleapis.com/model-cards/documents/gemini-2.5-pro-preview.pdf}
\showURL{%
\tempurl}


\bibitem[Gordon et~al\mbox{.}(2022)]%
        {gordon_jury_2022}
\bibfield{author}{\bibinfo{person}{Mitchell~L. Gordon}, \bibinfo{person}{Michelle~S. Lam}, \bibinfo{person}{Joon~Sung Park}, \bibinfo{person}{Kayur Patel}, \bibinfo{person}{Jeff Hancock}, \bibinfo{person}{Tatsunori Hashimoto}, {and} \bibinfo{person}{Michael~S. Bernstein}.} \bibinfo{year}{2022}\natexlab{}.
\newblock \showarticletitle{Jury {Learning}: {Integrating} {Dissenting} {Voices} into {Machine} {Learning} {Models}}. In \bibinfo{booktitle}{\emph{{CHI} {Conference} on {Human} {Factors} in {Computing} {Systems}}}. \bibinfo{publisher}{ACM}, \bibinfo{address}{New Orleans LA USA}, \bibinfo{pages}{1--19}.
\newblock
\showISBNx{978-1-4503-9157-3}
\href{https://doi.org/10.1145/3491102.3502004}{doi:\nolinkurl{10.1145/3491102.3502004}}


\bibitem[Iqbal et~al\mbox{.}(2024)]%
        {iqbal_openfactcheck_2024}
\bibfield{author}{\bibinfo{person}{Hasan Iqbal}, \bibinfo{person}{Yuxia Wang}, \bibinfo{person}{Minghan Wang}, \bibinfo{person}{Georgi~Nenkov Georgiev}, \bibinfo{person}{Jiahui Geng}, \bibinfo{person}{Iryna Gurevych}, {and} \bibinfo{person}{Preslav Nakov}.} \bibinfo{year}{2024}\natexlab{}.
\newblock \showarticletitle{{OpenFactCheck}: {A} {Unified} {Framework} for {Factuality} {Evaluation} of {LLMs}}. In \bibinfo{booktitle}{\emph{Proceedings of the 2024 {Conference} on {Empirical} {Methods} in {Natural} {Language} {Processing}: {System} {Demonstrations}}}. \bibinfo{publisher}{Association for Computational Linguistics}, \bibinfo{address}{Miami, Florida, USA}, \bibinfo{pages}{219--229}.
\newblock
\href{https://doi.org/10.18653/v1/2024.emnlp-demo.23}{doi:\nolinkurl{10.18653/v1/2024.emnlp-demo.23}}


\bibitem[{Jeremy Greer} et~al\mbox{.}(2025)]%
        {jeremy_greer_introducing_2025}
\bibfield{author}{\bibinfo{person}{{Jeremy Greer}}, \bibinfo{person}{{Manos Koukoumidis}}, \bibinfo{person}{{Konstantinos Aisopos}}, {and} \bibinfo{person}{{Michael Schuler}}.} \bibinfo{year}{2025}\natexlab{}.
\newblock \bibinfo{title}{Introducing {HallOumi}: {A} {State}-of-the-{Art} {Claim}-{Verification} {Model}}.
\newblock
\urldef\tempurl%
\url{https://oumi.ai/blog/posts/introducing-halloumi}
\showURL{%
\tempurl}


\bibitem[Kim et~al\mbox{.}(2024)]%
        {kim_prometheus_2024}
\bibfield{author}{\bibinfo{person}{Seungone Kim}, \bibinfo{person}{Jamin Shin}, \bibinfo{person}{Yejin Cho}, \bibinfo{person}{Joel Jang}, \bibinfo{person}{Shayne Longpre}, \bibinfo{person}{Hwaran Lee}, \bibinfo{person}{Sangdoo Yun}, \bibinfo{person}{Seongjin Shin}, \bibinfo{person}{Sungdong Kim}, \bibinfo{person}{James Thorne}, {and} \bibinfo{person}{Minjoon Seo}.} \bibinfo{year}{2024}\natexlab{}.
\newblock \bibinfo{title}{Prometheus: {Inducing} {Fine}-grained {Evaluation} {Capability} in {Language} {Models}}.
\newblock
\href{https://doi.org/10.48550/arXiv.2310.08491}{doi:\nolinkurl{10.48550/arXiv.2310.08491}}
\newblock
\shownote{arXiv:2310.08491 [cs]}.


\bibitem[Li et~al\mbox{.}(2023)]%
        {li_halueval_2023}
\bibfield{author}{\bibinfo{person}{Junyi Li}, \bibinfo{person}{Xiaoxue Cheng}, \bibinfo{person}{Wayne~Xin Zhao}, \bibinfo{person}{Jian-Yun Nie}, {and} \bibinfo{person}{Ji-Rong Wen}.} \bibinfo{year}{2023}\natexlab{}.
\newblock \bibinfo{title}{{HaluEval}: {A} {Large}-{Scale} {Hallucination} {Evaluation} {Benchmark} for {Large} {Language} {Models}}.
\newblock
\href{https://doi.org/10.48550/arXiv.2305.11747}{doi:\nolinkurl{10.48550/arXiv.2305.11747}}
\newblock
\shownote{arXiv:2305.11747}.


\bibitem[Lin et~al\mbox{.}(2021)]%
        {lin_truthfulqa_2021}
\bibfield{author}{\bibinfo{person}{Stephanie Lin}, \bibinfo{person}{Jacob Hilton}, {and} \bibinfo{person}{Owain Evans}.} \bibinfo{year}{2021}\natexlab{}.
\newblock \bibinfo{title}{{TruthfulQA}: {Measuring} {How} {Models} {Mimic} {Human} {Falsehoods}}.
\newblock
\urldef\tempurl%
\url{https://arxiv.org/abs/2109.07958v2}
\showURL{%
\tempurl}


\bibitem[Liu et~al\mbox{.}(2023)]%
        {liu_g-eval_2023}
\bibfield{author}{\bibinfo{person}{Yang Liu}, \bibinfo{person}{Dan Iter}, \bibinfo{person}{Yichong Xu}, \bibinfo{person}{Shuohang Wang}, \bibinfo{person}{Ruochen Xu}, {and} \bibinfo{person}{Chenguang Zhu}.} \bibinfo{year}{2023}\natexlab{}.
\newblock \showarticletitle{G-{Eval}: {NLG} {Evaluation} using {Gpt}-4 with {Better} {Human} {Alignment}}. In \bibinfo{booktitle}{\emph{Proceedings of the 2023 {Conference} on {Empirical} {Methods} in {Natural} {Language} {Processing}}}. \bibinfo{publisher}{Association for Computational Linguistics}, \bibinfo{address}{Singapore}, \bibinfo{pages}{2511--2522}.
\newblock
\href{https://doi.org/10.18653/v1/2023.emnlp-main.153}{doi:\nolinkurl{10.18653/v1/2023.emnlp-main.153}}


\bibitem[Luo et~al\mbox{.}(2024)]%
        {luo_halludial_2024}
\bibfield{author}{\bibinfo{person}{Wen Luo}, \bibinfo{person}{Tianshu Shen}, \bibinfo{person}{Wei Li}, \bibinfo{person}{Guangyue Peng}, \bibinfo{person}{Richeng Xuan}, \bibinfo{person}{Houfeng Wang}, {and} \bibinfo{person}{Xi Yang}.} \bibinfo{year}{2024}\natexlab{}.
\newblock \bibinfo{title}{{HalluDial}: {A} {Large}-{Scale} {Benchmark} for {Automatic} {Dialogue}-{Level} {Hallucination} {Evaluation}}.
\newblock
\href{https://doi.org/10.48550/arXiv.2406.07070}{doi:\nolinkurl{10.48550/arXiv.2406.07070}}
\newblock
\shownote{arXiv:2406.07070 version: 1}.


\bibitem[{Meta}(2025)]%
        {meta_llama_2025}
\bibfield{author}{\bibinfo{person}{{Meta}}.} \bibinfo{year}{2025}\natexlab{}.
\newblock \bibinfo{title}{Llama 4}.
\newblock
\urldef\tempurl%
\url{https://github.com/meta-llama/llama-models/blob/main/models/llama4/MODEL_CARD.md}
\showURL{%
\tempurl}


\bibitem[Metropolitansky and Larson(2025)]%
        {metropolitansky_towards_2025}
\bibfield{author}{\bibinfo{person}{Dasha Metropolitansky} {and} \bibinfo{person}{Jonathan Larson}.} \bibinfo{year}{2025}\natexlab{}.
\newblock \bibinfo{title}{Towards {Effective} {Extraction} and {Evaluation} of {Factual} {Claims}}.
\newblock
\href{https://doi.org/10.48550/arXiv.2502.10855}{doi:\nolinkurl{10.48550/arXiv.2502.10855}}
\newblock
\shownote{arXiv:2502.10855 [cs]}.


\bibitem[Narayan et~al\mbox{.}(2018)]%
        {narayan_dont_2018}
\bibfield{author}{\bibinfo{person}{Shashi Narayan}, \bibinfo{person}{Shay~B. Cohen}, {and} \bibinfo{person}{Mirella Lapata}.} \bibinfo{year}{2018}\natexlab{}.
\newblock \showarticletitle{Don’t {Give} {Me} the {Details}, {Just} the {Summary}! {Topic}-{Aware} {Convolutional} {Neural} {Networks} for {Extreme} {Summarization}}. In \bibinfo{booktitle}{\emph{Proceedings of the 2018 {Conference} on {Empirical} {Methods} in {Natural} {Language} {Processing}}}. \bibinfo{publisher}{Association for Computational Linguistics}, \bibinfo{address}{Brussels, Belgium}, \bibinfo{pages}{1797--1807}.
\newblock
\href{https://doi.org/10.18653/v1/D18-1206}{doi:\nolinkurl{10.18653/v1/D18-1206}}


\bibitem[{OpenAI}(2024a)]%
        {openai_gpt-41_2024}
\bibfield{author}{\bibinfo{person}{{OpenAI}}.} \bibinfo{year}{2024}\natexlab{a}.
\newblock \bibinfo{title}{{GPT}-4.1 nano}.
\newblock
\urldef\tempurl%
\url{https://platform.openai.com/docs/models/gpt-4.1-nano}
\showURL{%
\tempurl}


\bibitem[{OpenAI}(2024b)]%
        {openai_gpt-4o_2024-1}
\bibfield{author}{\bibinfo{person}{{OpenAI}}.} \bibinfo{year}{2024}\natexlab{b}.
\newblock \bibinfo{title}{{GPT}-4o mini: advancing cost-efficient intelligence}.
\newblock
\urldef\tempurl%
\url{https://openai.com/index/gpt-4o-mini-advancing-cost-efficient-intelligence/}
\showURL{%
\tempurl}


\bibitem[{OpenAI}(2024c)]%
        {openai_gpt-4o_2024}
\bibfield{author}{\bibinfo{person}{{OpenAI}}.} \bibinfo{year}{2024}\natexlab{c}.
\newblock \bibinfo{title}{{GPT}-4o {System} {Card}}.
\newblock
\urldef\tempurl%
\url{https://cdn.openai.com/gpt-4o-system-card.pdf}
\showURL{%
\tempurl}


\bibitem[{OpenAI}(2024d)]%
        {openai_openai_2024}
\bibfield{author}{\bibinfo{person}{{OpenAI}}.} \bibinfo{year}{2024}\natexlab{d}.
\newblock \bibinfo{title}{{OpenAI} o1 {System} {Card}}.
\newblock
\urldef\tempurl%
\url{https://cdn.openai.com/o1-system-card-20241205.pdf}
\showURL{%
\tempurl}


\bibitem[{OpenAI}(2025a)]%
        {openai_gpt-41_2025-1}
\bibfield{author}{\bibinfo{person}{{OpenAI}}.} \bibinfo{year}{2025}\natexlab{a}.
\newblock \bibinfo{title}{{GPT}-4.1}.
\newblock
\urldef\tempurl%
\url{https://platform.openai.com/docs/models/gpt-4.1}
\showURL{%
\tempurl}


\bibitem[{OpenAI}(2025b)]%
        {openai_gpt-41_2025}
\bibfield{author}{\bibinfo{person}{{OpenAI}}.} \bibinfo{year}{2025}\natexlab{b}.
\newblock \bibinfo{title}{{GPT}-4.1 mini}.
\newblock
\urldef\tempurl%
\url{https://platform.openai.com/docs/models/gpt-4.1-mini}
\showURL{%
\tempurl}


\bibitem[{OpenAI}(2025c)]%
        {openai_openai_2025}
\bibfield{author}{\bibinfo{person}{{OpenAI}}.} \bibinfo{year}{2025}\natexlab{c}.
\newblock \bibinfo{title}{{OpenAI} o3 and o4-mini {System} {Card}}.
\newblock
\urldef\tempurl%
\url{https://cdn.openai.com/pdf/2221c875-02dc-4789-800b-e7758f3722c1/o3-and-o4-mini-system-card.pdf}
\showURL{%
\tempurl}


\bibitem[{Oumi Community}(2025)]%
        {oumi_community_oumi_2025}
\bibfield{author}{\bibinfo{person}{{Oumi Community}}.} \bibinfo{year}{2025}\natexlab{}.
\newblock \bibinfo{title}{Oumi: an {Open}, {End}-to-end {Platform} for {Building} {Large} {Foundation} {Models}}.
\newblock
\urldef\tempurl%
\url{https://github.com/oumi-ai/oumi}
\showURL{%
\tempurl}


\bibitem[{Panos Achlioptas} et~al\mbox{.}(2025)]%
        {panos_achlioptas_halloumi-8b-classifier_2025}
\bibfield{author}{\bibinfo{person}{{Panos Achlioptas}}, \bibinfo{person}{{Jeremy Greer}}, \bibinfo{person}{{Konstantinos Aisopos}}, \bibinfo{person}{{Michael Schuler}}, \bibinfo{person}{{Oussama Elachqar}}, {and} \bibinfo{person}{{Emmanouil Koukoumidis}}.} \bibinfo{year}{2025}\natexlab{}.
\newblock \bibinfo{title}{{HallOumi}-{8B}-classifier}.
\newblock
\urldef\tempurl%
\url{https://huggingface.co/oumi-ai/HallOumi-8B-classifier}
\showURL{%
\tempurl}


\bibitem[Schaekermann et~al\mbox{.}(2020)]%
        {schaekermann_ambiguity-aware_2020}
\bibfield{author}{\bibinfo{person}{Mike Schaekermann}, \bibinfo{person}{Graeme Beaton}, \bibinfo{person}{Elaheh Sanoubari}, \bibinfo{person}{Andrew Lim}, \bibinfo{person}{Kate Larson}, {and} \bibinfo{person}{Edith Law}.} \bibinfo{year}{2020}\natexlab{}.
\newblock \showarticletitle{Ambiguity-aware {AI} {Assistants} for {Medical} {Data} {Analysis}}. In \bibinfo{booktitle}{\emph{Proceedings of the 2020 {CHI} {Conference} on {Human} {Factors} in {Computing} {Systems}}}. \bibinfo{publisher}{ACM}, \bibinfo{address}{Honolulu HI USA}, \bibinfo{pages}{1--14}.
\newblock
\showISBNx{978-1-4503-6708-0}
\href{https://doi.org/10.1145/3313831.3376506}{doi:\nolinkurl{10.1145/3313831.3376506}}


\bibitem[Swayamdipta et~al\mbox{.}(2020)]%
        {swayamdipta_dataset_2020}
\bibfield{author}{\bibinfo{person}{Swabha Swayamdipta}, \bibinfo{person}{Roy Schwartz}, \bibinfo{person}{Nicholas Lourie}, \bibinfo{person}{Yizhong Wang}, \bibinfo{person}{Hannaneh Hajishirzi}, \bibinfo{person}{Noah~A. Smith}, {and} \bibinfo{person}{Yejin Choi}.} \bibinfo{year}{2020}\natexlab{}.
\newblock \showarticletitle{Dataset {Cartography}: {Mapping} and {Diagnosing} {Datasets} with {Training} {Dynamics}}. In \bibinfo{booktitle}{\emph{Proceedings of the 2020 {Conference} on {Empirical} {Methods} in {Natural} {Language} {Processing} ({EMNLP})}}. \bibinfo{publisher}{Association for Computational Linguistics}, \bibinfo{address}{Online}, \bibinfo{pages}{9275--9293}.
\newblock
\href{https://doi.org/10.18653/v1/2020.emnlp-main.746}{doi:\nolinkurl{10.18653/v1/2020.emnlp-main.746}}


\bibitem[Szabó(2020)]%
        {szabo_compositionality_2020}
\bibfield{author}{\bibinfo{person}{Zoltán~Gendler Szabó}.} \bibinfo{year}{2020}\natexlab{}.
\newblock \bibinfo{title}{Compositionality}.
\newblock
\urldef\tempurl%
\url{https://plato.stanford.edu/archives/fall2024/entries/compositionality/}
\showURL{%
\tempurl}


\bibitem[Tang et~al\mbox{.}(2024)]%
        {tang_minicheck_2024}
\bibfield{author}{\bibinfo{person}{Liyan Tang}, \bibinfo{person}{Philippe Laban}, {and} \bibinfo{person}{Greg Durrett}.} \bibinfo{year}{2024}\natexlab{}.
\newblock \showarticletitle{{MiniCheck}: {Efficient} {Fact}-{Checking} of {LLMs} on {Grounding} {Documents}}. In \bibinfo{booktitle}{\emph{Proceedings of the 2024 {Conference} on {Empirical} {Methods} in {Natural} {Language} {Processing}}}. \bibinfo{publisher}{Association for Computational Linguistics}, \bibinfo{address}{Miami, Florida, USA}, \bibinfo{pages}{8818--8847}.
\newblock
\href{https://doi.org/10.18653/v1/2024.emnlp-main.499}{doi:\nolinkurl{10.18653/v1/2024.emnlp-main.499}}


\bibitem[Tang et~al\mbox{.}(2021)]%
        {tang_confit_2021}
\bibfield{author}{\bibinfo{person}{Xiangru Tang}, \bibinfo{person}{Arjun Nair}, \bibinfo{person}{Borui Wang}, \bibinfo{person}{Bingyao Wang}, \bibinfo{person}{Jai Desai}, \bibinfo{person}{Aaron Wade}, \bibinfo{person}{Haoran Li}, \bibinfo{person}{Asli Celikyilmaz}, \bibinfo{person}{Yashar Mehdad}, {and} \bibinfo{person}{Dragomir Radev}.} \bibinfo{year}{2021}\natexlab{}.
\newblock \bibinfo{title}{{CONFIT}: {Toward} {Faithful} {Dialogue} {Summarization} with {Linguistically}-{Informed} {Contrastive} {Fine}-tuning}.
\newblock
\href{https://doi.org/10.18653/v1/2022.naacl-main.415}{doi:\nolinkurl{10.18653/v1/2022.naacl-main.415}}


\bibitem[Wang et~al\mbox{.}(2020)]%
        {wang_asking_2020}
\bibfield{author}{\bibinfo{person}{Alex Wang}, \bibinfo{person}{Kyunghyun Cho}, {and} \bibinfo{person}{Mike Lewis}.} \bibinfo{year}{2020}\natexlab{}.
\newblock \showarticletitle{Asking and {Answering} {Questions} to {Evaluate} the {Factual} {Consistency} of {Summaries}}. In \bibinfo{booktitle}{\emph{Proceedings of the 58th {Annual} {Meeting} of the {Association} for {Computational} {Linguistics}}}. \bibinfo{publisher}{Association for Computational Linguistics}, \bibinfo{address}{Online}, \bibinfo{pages}{5008--5020}.
\newblock
\href{https://doi.org/10.18653/v1/2020.acl-main.450}{doi:\nolinkurl{10.18653/v1/2020.acl-main.450}}


\bibitem[Wang et~al\mbox{.}(2022)]%
        {wang_analyzing_2022}
\bibfield{author}{\bibinfo{person}{Bin Wang}, \bibinfo{person}{Chen Zhang}, \bibinfo{person}{Yan Zhang}, \bibinfo{person}{Yiming Chen}, {and} \bibinfo{person}{Haizhou Li}.} \bibinfo{year}{2022}\natexlab{}.
\newblock \bibinfo{title}{Analyzing and {Evaluating} {Faithfulness} in {Dialogue} {Summarization}}.
\newblock
\href{https://doi.org/10.48550/arXiv.2210.11777}{doi:\nolinkurl{10.48550/arXiv.2210.11777}}
\newblock
\shownote{arXiv:2210.11777}.


\bibitem[Wei et~al\mbox{.}(2025)]%
        {wei_browsecomp_2025}
\bibfield{author}{\bibinfo{person}{Jason Wei}, \bibinfo{person}{Zhiqing Sun}, \bibinfo{person}{Spencer Papay}, \bibinfo{person}{Scott McKinney}, \bibinfo{person}{Jeffrey Han}, \bibinfo{person}{Isa Fulford}, \bibinfo{person}{Hyung~Won Chung}, \bibinfo{person}{Alex~Tachard Passos}, \bibinfo{person}{William Fedus}, {and} \bibinfo{person}{Amelia Glaese}.} \bibinfo{year}{2025}\natexlab{}.
\newblock \bibinfo{title}{{BrowseComp}: {A} {Simple} {Yet} {Challenging} {Benchmark} for {Browsing} {Agents}}.
\newblock
\href{https://doi.org/10.48550/arXiv.2504.12516}{doi:\nolinkurl{10.48550/arXiv.2504.12516}}
\newblock
\shownote{arXiv:2504.12516}.


\bibitem[Wei et~al\mbox{.}(2023)]%
        {wei_chain--thought_2023}
\bibfield{author}{\bibinfo{person}{Jason Wei}, \bibinfo{person}{Xuezhi Wang}, \bibinfo{person}{Dale Schuurmans}, \bibinfo{person}{Maarten Bosma}, \bibinfo{person}{Brian Ichter}, \bibinfo{person}{Fei Xia}, \bibinfo{person}{Ed Chi}, \bibinfo{person}{Quoc Le}, {and} \bibinfo{person}{Denny Zhou}.} \bibinfo{year}{2023}\natexlab{}.
\newblock \bibinfo{title}{Chain-of-{Thought} {Prompting} {Elicits} {Reasoning} in {Large} {Language} {Models}}.
\newblock
\href{https://doi.org/10.48550/arXiv.2201.11903}{doi:\nolinkurl{10.48550/arXiv.2201.11903}}
\newblock
\shownote{arXiv:2201.11903 [cs]}.


\bibitem[Zhang et~al\mbox{.}(2025)]%
        {zhang_survey_2025}
\bibfield{author}{\bibinfo{person}{Qiyuan Zhang}, \bibinfo{person}{Fuyuan Lyu}, \bibinfo{person}{Zexu Sun}, \bibinfo{person}{Lei Wang}, \bibinfo{person}{Weixu Zhang}, \bibinfo{person}{Wenyue Hua}, \bibinfo{person}{Haolun Wu}, \bibinfo{person}{Zhihan Guo}, \bibinfo{person}{Yufei Wang}, \bibinfo{person}{Niklas Muennighoff}, \bibinfo{person}{Irwin King}, \bibinfo{person}{Xue Liu}, {and} \bibinfo{person}{Chen Ma}.} \bibinfo{year}{2025}\natexlab{}.
\newblock \bibinfo{title}{A {Survey} on {Test}-{Time} {Scaling} in {Large} {Language} {Models}: {What}, {How}, {Where}, and {How} {Well}?}
\newblock
\href{https://doi.org/10.48550/arXiv.2503.24235}{doi:\nolinkurl{10.48550/arXiv.2503.24235}}
\newblock
\shownote{arXiv:2503.24235}.


\bibitem[Zhang et~al\mbox{.}(2023)]%
        {zhang_taxonomy_2023}
\bibfield{author}{\bibinfo{person}{Wenbo Zhang}, \bibinfo{person}{Hangzhi Guo}, \bibinfo{person}{Ian~D. Kivlichan}, \bibinfo{person}{Vinodkumar Prabhakaran}, \bibinfo{person}{Davis Yadav}, {and} \bibinfo{person}{Amulya Yadav}.} \bibinfo{year}{2023}\natexlab{}.
\newblock \bibinfo{title}{A {Taxonomy} of {Rater} {Disagreements}: {Surveying} {Challenges} \& {Opportunities} from the {Perspective} of {Annotating} {Online} {Toxicity}}.
\newblock
\href{https://doi.org/10.48550/arXiv.2311.04345}{doi:\nolinkurl{10.48550/arXiv.2311.04345}}
\newblock
\shownote{arXiv:2311.04345 [cs]}.


\bibitem[Zheng et~al\mbox{.}(2023)]%
        {zheng_judging_2023}
\bibfield{author}{\bibinfo{person}{Lianmin Zheng}, \bibinfo{person}{Wei-Lin Chiang}, \bibinfo{person}{Ying Sheng}, \bibinfo{person}{Siyuan Zhuang}, \bibinfo{person}{Zhanghao Wu}, \bibinfo{person}{Yonghao Zhuang}, \bibinfo{person}{Zi Lin}, \bibinfo{person}{Zhuohan Li}, \bibinfo{person}{Dacheng Li}, \bibinfo{person}{Eric~P. Xing}, \bibinfo{person}{Hao Zhang}, \bibinfo{person}{Joseph~E. Gonzalez}, {and} \bibinfo{person}{Ion Stoica}.} \bibinfo{year}{2023}\natexlab{}.
\newblock \bibinfo{title}{Judging {LLM}-as-a-{Judge} with {MT}-{Bench} and {Chatbot} {Arena}}.
\newblock
\href{https://doi.org/10.48550/arXiv.2306.05685}{doi:\nolinkurl{10.48550/arXiv.2306.05685}}
\newblock
\shownote{arXiv:2306.05685 [cs]}.


\end{thebibliography}

\appendix

\newpage
\section{Hallucination example}

\begin{figure}[h]
    \centering
    \includegraphics[width=\linewidth]{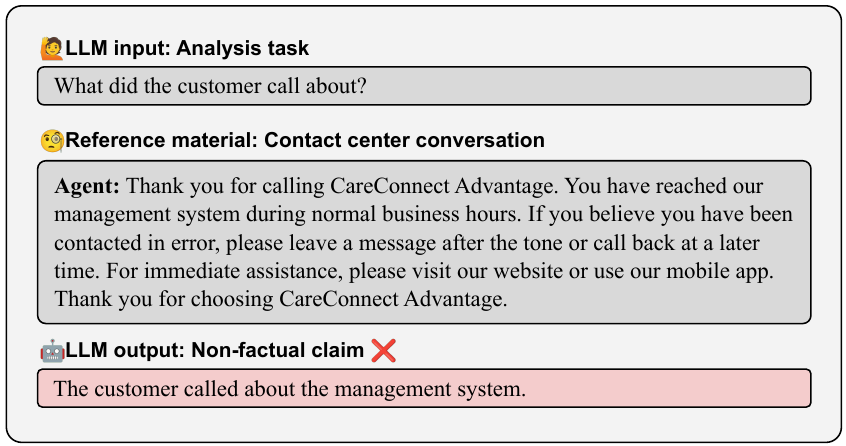}
    \caption{An example of a typical LLM hallucination observed when automatically analyzing contact center conversations using LLMs. In this example, the LLM is asked ``What did the customer call about?'' in the conversation only contained a voicemail. 
    The LLM wrongly responded that \textit{The customer called about the management system}, presumably because such voicemail conversations---common in contact center scenarios---are out of distribution of many LLMs' training data. 
    }
    \label{fig:hallucination}
\end{figure}

\section{Prompts}\label{sec:appendix-prompts}

\subsection{\texttt{3D\_WITH\_TTC}}\label{sec:3d-with-ttc}
\begin{small}
\begin{lstlisting}
3D_FACTUALITY_SYSTEM_PROMPT_WITH_TTC = """Given a conversation and a short answer, verify the short answer by referencing the conversation. First, break down the short answer into claims using the `A Step to Extract Claims` below. Next, verify each part of the claim and the relation between each part of the claim using `Steps to Evaluate Each Claim` below.

## A Step to Extract Claims ##
Step 1: Identify claims from the short answer. Example: "Customer was annoyed about slow delivery" -> "There was a delivery", "The delivery was slow", "Customer was annoyed", "Customer was annoyed specifically about slow delivery"

## Steps to Evaluate Each Claim ##
Step 2: In each claim, identify words that have concrete meanings. Example: "There was a delivery" -> "delivery". Verify those words by finding explicit mentions or references. When a word or a phrase can be interpreted in more than one way, see if at least one interpretation can be verified. Example: If a conversation includes discussions of receiving email notifications, this verifies one meaning of "delivery".
Step 3: In each claim, identify words that subjectively describe other words having concrete meanings. These words often describe a product or a service. Example: "The delivery was slow" -> "slow". Verify these words loosely with the context of the conversation.
Step 4: In each claim, identify words that are about subjective interpretation of the conversation. These words often describe sentiments and emotions from a third-person point of view. Example: "Customer was annoyed" -> "annoyed". Verify these words by finding minimal implicit evidence. Example: "annoyed" is verified with implicit evidence reflecting negative sentiment. 
Step 5: In each claim, verify the relation between words. Focus on verifying the relation between words, while ignoring the verifications of the words themselves in this step. Verify the relation with explicit evidence or by inferring the reason behind an action or a message. Example: "Customer was annoyed specifically about slow delivery" -> Verify that the source of a customer's sentiment was indeed the "slow delivery" while ignoring the verifications of "slow" and "annoyed". If a customer asks about filing a complaint after discussing slow delivery without explicitly expressing a negative sentiment, the customer must have been annoyed by the slow delivery. This inferred reason behind the customer's action verifies the relation.

## Output Format as JSON ##
claims: list of all the claims generated above in the mentioned format.
reasoning: A concise summary of the reasoning for the final answer.
answer: True or False (True if short_answer is verified; otherwise, False)."""
\end{lstlisting}
\end{small}

\subsection{\texttt{3D\_NO\_TTC}}\label{sec:3d-no-ttc}
\begin{small}
\begin{lstlisting}
3D_FACTUALITY_SYSTEM_PROMPT_NO_TTC = """Given a conversation and a short answer, verify the short answer by referencing the conversation. First, break down the short answer into claims using the `A Step to Extract Claims` below. Next, verify each part of the claim and the relation between each part of the claim using `Steps to Evaluate Each Claim` below.

## A Step to Extract Claims ##
Step 1: Identify claims from the short answer. Example: "Customer was annoyed about slow delivery" -> "There was a delivery", "The delivery was slow", "Customer was annoyed", "Customer was annoyed specifically about slow delivery"

## Steps to Evaluate Each Claim ##
Step 2: In each claim, identify words that have concrete meanings. Example: "There was a delivery" -> "delivery". Verify those words by finding explicit mentions or references. When a word or a phrase can be interpreted in more than one way, see if at least one interpretation can be verified. Example: If a conversation includes discussions of receiving email notifications, this verifies one meaning of "delivery".
Step 3: In each claim, identify words that subjectively describe other words having concrete meanings. These words often describe a product or a service. Example: "The delivery was slow" -> "slow". Verify these words loosely with the context of the conversation.
Step 4: In each claim, identify words that are about subjective interpretation of the conversation. These words often describe sentiments and emotions from a third-person point of view. Example: "Customer was annoyed" -> "annoyed". Verify these words by finding minimal implicit evidence. Example: "annoyed" is verified with implicit evidence reflecting negative sentiment. 
Step 5: In each claim, verify the relation between words. Focus on verifying the relation between words, while ignoring the verifications of the words themselves in this step. Verify the relation with explicit evidence or by inferring the reason behind an action or a message. Example: "Customer was annoyed specifically about slow delivery" -> Verify that the source of a customer's sentiment was indeed the "slow delivery" while ignoring the verifications of "slow" and "annoyed". If a customer asks about filing a complaint after discussing slow delivery without explicitly expressing a negative sentiment, the customer must have been annoyed by the slow delivery. This inferred reason behind the customer's action verifies the relation.

## Output Format as JSON ##
answer: True or False (True if short_answer is verified; otherwise, False)."""
\end{lstlisting}
\end{small}

\subsection{\texttt{BASIC\_WITH\_TTC}}\label{sec:basic-with-ttc}
\begin{small}
\begin{lstlisting}
BASIC_FACTUALITY_SYSTEM_PROMPT_WITH_TTC = """Given a conversation and a claim about that conversation, determine if the claim is factual, i.e., supported by the conversation.

### Output Format as JSON:
reasoning: A concise summary of the reasoning for the final answer.
answer: True or False (True if the claim is factual; otherwise, False)."""
\end{lstlisting}
\end{small}

\subsection{\texttt{BASIC\_NO\_TTC}}\label{sec:basic-no-ttc}
\begin{small}
\begin{lstlisting}
BASIC_FACTUALITY_SYSTEM_PROMPT_NO_TTC = """Given a conversation and a claim about that conversation, determine if the claim is factual, i.e., supported by the conversation.

### Output Format as JSON:
answer: True or False (True if claim is factual; otherwise, False)."""
\end{lstlisting}
\end{small}

\subsection{User prompt}
\begin{small}
\begin{lstlisting}
FACTUALITY_USER_PROMPT = """### Conversation ###
{conversation}

### Short answer ###
{short_answer}"""
\end{lstlisting}
\end{small}

\subsection{Specifications of XML output format used with Anthropic models}
See \url{https://github.com/cresta/fect/blob/main/scripts/constants/prompts.py}.

\section{Models}

\begin{table}[h]
    \centering
    \footnotesize
    \begin{tabular}{c c c} \toprule
         \textbf{Model Name} & \textbf{Model ID} & \textbf{Source} \\ \midrule
         gemini-2.5-flash & gemini-2.5-flash-preview-04-17 & \cite{google_gemini_2025-1}\\
        gemini-2.5-pro& gemini-2.5-pro-preview-05-06 & \cite{google_gemini_2025} \\
        o1& o1-2024-12-17 & \cite{openai_openai_2024}\\
        o3& o3-2025-04-16 & \cite{openai_openai_2025}\\
        o4-mini& o4-mini-2025-04-16 & \cite{openai_openai_2025} \\
        claude-sonnet-3.5 & anthropic.claude-3-5-sonnet-20240620-v1:0 & \cite{anthropic_claude_2024}\\
        claude-sonnet-3.7 & anthropic.claude-3-7-sonnet-20250219-v1:0 & \cite{anthropic_claude_2025} \\
        claude-sonnet-4& anthropic.claude-sonnet-4-20250514-v1:0 & \cite{anthropic_system_2025} \\
        deepseek-r1& deepseek-r1-basic & \cite{deepseek_deepseek-r1_2025} \\
        gpt-4.1& gpt-4.1-2025-04-14 & \cite{openai_gpt-41_2025-1} \\
        gpt-4.1-mini& gpt-4.1-mini-2025-04-14 & \cite{openai_gpt-41_2025}\\
        gpt-4.1-nano& gpt-4.1-nano-2025-04-14 & \cite{openai_gpt-41_2024} \\
        gpt-4o& gpt-4o-2024-08-06 & \cite{openai_gpt-4o_2024-1} \\
        gpt-4o-mini& gpt-4o-mini-2024-07-18 & \cite{openai_gpt-4o_2024}\\
        llama4-maverick& llama4-maverick-instruct-basic & \cite{meta_llama_2025} \\ 
        HallOumi-8B & HallOumi-8B & \cite{oumi_community_oumi_2025}\\
        HallOumi-8B-Classifier & HallOumi-8B-Classifier & \cite{panos_achlioptas_halloumi-8b-classifier_2025}\\ 
        \bottomrule
    \end{tabular}
    \caption{Names, IDs and sources of the models tested in our ablation study.}
    \label{tab:list-models}
\end{table}

\newpage
\section{Ablation results}

\begin{figure}[h!]
    \centering
    \includegraphics[width=\linewidth]{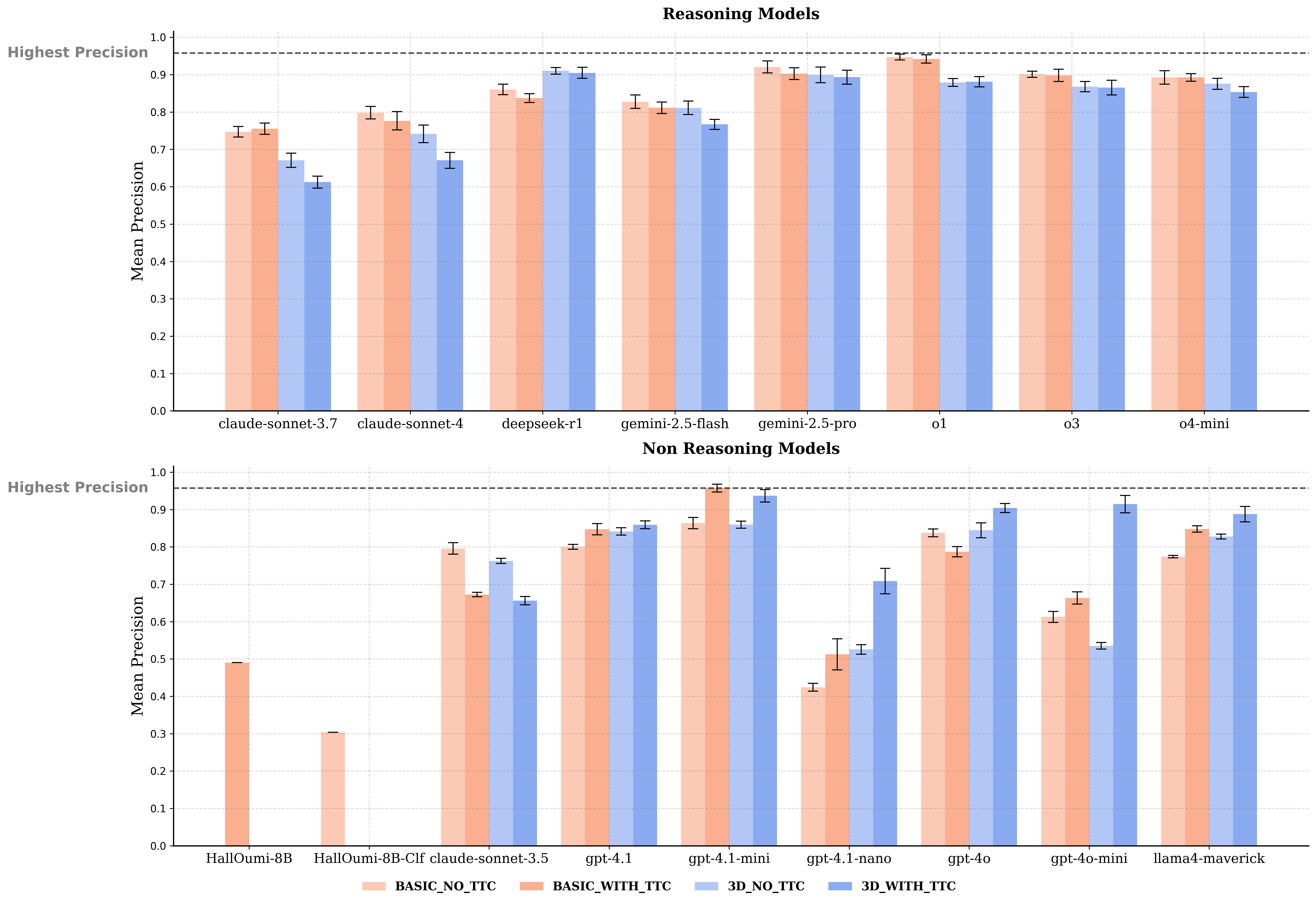}
    \caption{Means and 95\% CIs of precisions achieved with reasoning models and non-reasoning models and with four prompts. Dashed horizontal lines indicate the highest precision of 0.96 achieved with GPT-4.1-mini with the \texttt{BASIC\_WITH\_TTC} prompt.}
    \label{fig:precision}
\end{figure}

\begin{figure}[h!]
    \centering
    \includegraphics[width=\linewidth]{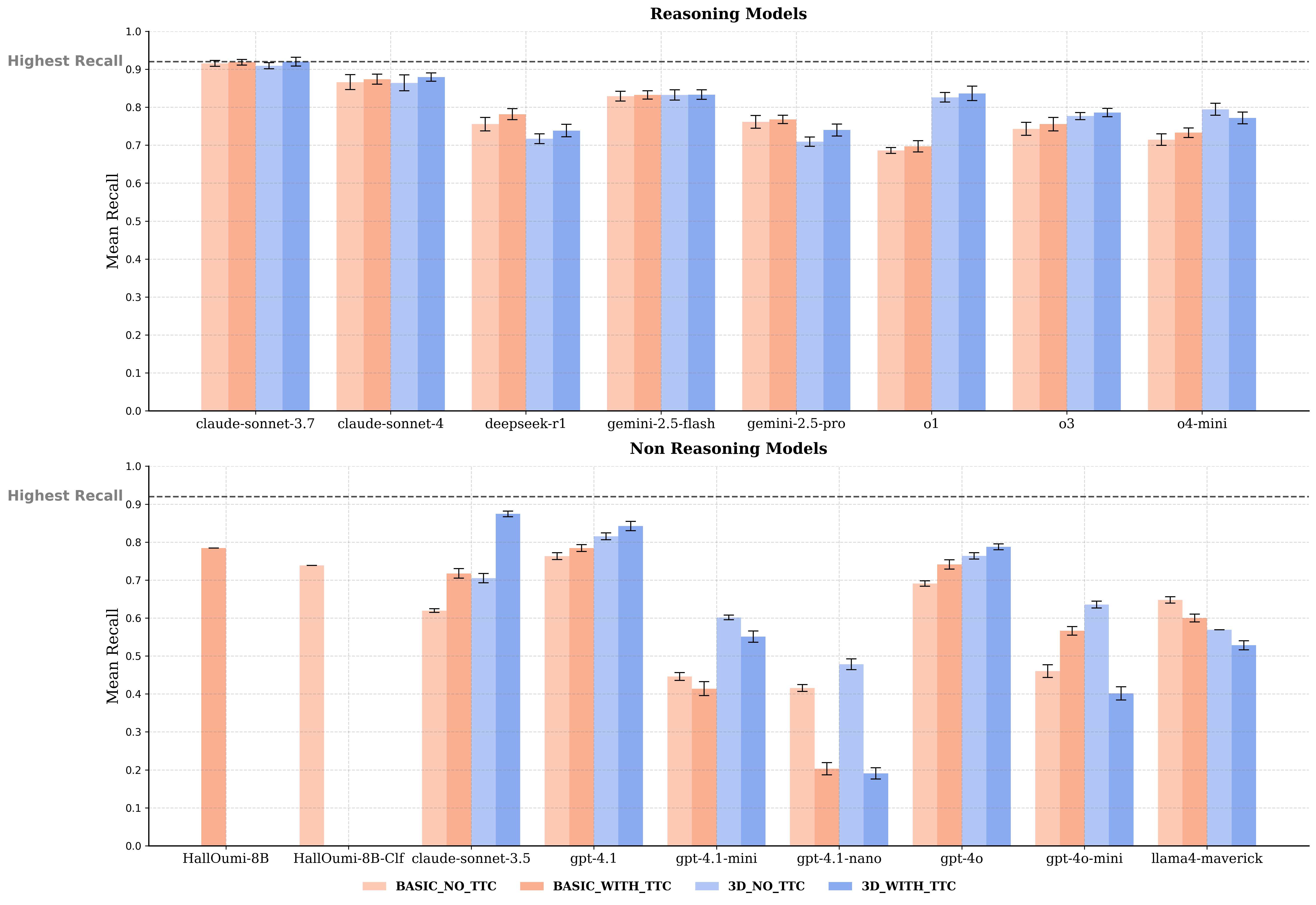}
    \caption{Means and 95\% CIs of recalls achieved with reasoning models and non-reasoning models and with four prompts. Dashed horizontal lines indicate the highest recall of 0.92 achieved with Claude-sonnet-3.7 with the \texttt{3D\_WITH\_TTC} prompt.}
    \label{fig:recall}
\end{figure}

\newpage

\begin{table}[h!]
\centering
\begin{tabular}{lp{0.8cm}p{0.8cm}p{0.8cm}p{0.8cm}}
\toprule
 & \multicolumn{4}{c}{\textbf{Precision scores (mean ± std)}} \\
 \textbf{Model Name} &
\texttt{BASIC\_} \texttt{NO\_} \texttt{TTC} & \texttt{BASIC\_} \texttt{WITH\_} \texttt{TTC} & \texttt{3D\_} \texttt{NO\_} \texttt{TTC} & \texttt{3D\_} \texttt{WITH\_} \texttt{TTC} \\
\midrule
claude-sonnet-3.5 & 0.80 & 0.67  & 0.76 & 0.65 \\
 & ± 0.02 &  ± 0.01 &  ± 0.01 &  ± 0.02 \\
claude-sonnet-3.7 & 0.75 \space\space  ± 0.02 & 0.76 \space\space ± 0.02 & 0.67 \space\space ± 0.03 & 0.61 \space\space ± 0.02 \\
claude-sonnet-4 & 0.80 \space\space ± 0.02 & 0.78 \space\space ± 0.02 & 0.74 \space\space ± 0.03 & 0.67 \space\space ± 0.02 \\
deepseek-r1 & 0.86 \space\space ± 0.02 & 0.84 \space\space ± 0.02 & \textbf{0.91} \space\space \textbf{± 0.01} & 0.90 \space\space ± 0.02 \\
gemini-2.5-flash & 0.83 \space\space ± 0.03 & 0.81 \space\space ± 0.02 & 0.81 \space\space ± 0.03 & 0.77 \space\space ± 0.02 \\
gemini-2.5-pro & 0.92 \space\space ± 0.02 & 0.90 \space\space ± 0.02 & 0.90 \space\space ± 0.03 & 0.89 \space\space ± 0.03 \\
gpt-4.1 & 0.80 \space\space ± 0.01 & 0.85 \space\space ± 0.02 & 0.84 \space\space ± 0.01 & 0.86 \space\space ± 0.02 \\
gpt-4.1-mini & 0.86 \space\space ± 0.02 & \textbf{0.96} \space\space \textbf{± 0.01} & 0.86 \space\space ± 0.01 & \textbf{0.94} \space\space \textbf{± 0.03} \\
gpt-4.1-nano & 0.42 \space\space ± 0.01 & 0.51 \space\space ± 0.06 & 0.53 \space\space ± 0.02 & 0.71 \space\space ± 0.08 \\
gpt-4o & 0.84 \space\space ± 0.02 & 0.79 \space\space ± 0.02 & 0.84 \space\space ± 0.03 & 0.91 \space\space± 0.02 \\
gpt-4o-mini & 0.61 \space\space± 0.02 & 0.66 \space\space± 0.02 & 0.54 \space\space± 0.01 & 0.92 \space\space± 0.03 \\
llama4-maverick & 0.77 \space\space± 0.00 & 0.85 \space\space± 0.01 & 0.83 \space\space± 0.01 & 0.92 \space\space ± 0.03 \\
o1 & \textbf{0.95} \space\space \textbf{± 0.01} & 0.94 \space\space ± 0.02 & 0.88 \space\space ± 0.01 & 0.88 \space\space ± 0.02 \\
o3 & 0.90 \space\space ± 0.01 & 0.90 \space\space ± 0.02 & 0.87 \space\space ± 0.02 & 0.87 \space\space ± 0.03 \\
o4-mini & 0.89 \space\space ± 0.03 & 0.89 \space\space ± 0.01 & 0.88 \space\space ± 0.02 & 0.85 \space\space ± 0.02 \\
HallOumi-8B & -- & 0.49 \space\space ± 0.00 & -- & -- \\
HallOumi-8B-Clf & 0.30 \space\space ± 0.00 & -- & -- & -- \\
\bottomrule
\end{tabular}
\caption{Mean and standard deviation of precision scores across 17 models under 4 prompting modes. Boldfaced scores indicate the best mean within each prompt mode.}
\label{tab:precision}
\end{table}

\newpage

\begin{table}[h!]
\centering
\begin{tabular}{lp{0.8cm}p{0.8cm}p{0.8cm}p{0.8cm}}
\toprule
 & \multicolumn{4}{c}{\textbf{Recall scores (mean ± std)}} \\
\textbf{Model Name} &
\texttt{BASIC\_} \texttt{NO\_} \texttt{TTC} & \texttt{BASIC\_} \texttt{WITH\_} \texttt{TTC} & \texttt{3D\_} \texttt{NO\_} \texttt{TTC} & \texttt{3D\_} \texttt{WITH\_} \texttt{TTC} \\ \midrule
claude-sonnet-3.5 & 0.62 \space\space ± 0.01 & 0.72 \space\space± 0.02 & 0.70 \space\space± 0.02 & 0.88 \space\space± 0.01 \\
claude-sonnet-3.7 & \textbf{0.92} \space\space\textbf{± 0.01}& \textbf{0.92} \space\space\textbf{± 0.01} & \textbf{0.91} \space\space \textbf{± 0.01} & \textbf{0.92 }\space\space\textbf{± 0.02} \\
claude-sonnet-4 & 0.87 \space\space± 0.02 & 0.87 \space\space± 0.01 & 0.86 \space\space± 0.02 & 0.88 \space\space± 0.01 \\
deepseek-r1 & 0.76 \space\space± 0.02 & 0.78 \space\space± 0.02 & 0.72 \space\space± 0.02 & 0.74 \space\space± 0.03 \\
gemini-2.5-flash & 0.83 \space\space± 0.02 & 0.83 \space\space± 0.02 & 0.83 \space\space± 0.02 & 0.83 \space\space± 0.02 \\
gemini-2.5-pro & 0.76 \space\space± 0.02 & 0.77 \space\space± 0.02 & 0.71 \space\space± 0.02 & 0.74 \space\space± 0.02 \\
gpt-4.1 & 0.76 \space\space ± 0.01 & 0.78 \space\space ± 0.01 & 0.82 \space\space ± 0.01 & 0.84 \space\space ± 0.02 \\
gpt-4.1-mini & 0.45 \space\space ± 0.01 & 0.41 \space\space ± 0.03 & 0.60 \space\space ± 0.01 & 0.55 \space\space ± 0.02 \\
gpt-4.1-nano & 0.42 \space\space ± 0.01 & 0.20 \space\space ± 0.02 & 0.48 \space\space ± 0.02 & 0.20 \space\space ± 0.03 \\
gpt-4o & 0.69 \space\space ± 0.01 & 0.74 \space\space ± 0.02 & 0.76 \space\space ± 0.01 & 0.79 \space\space ± 0.01 \\
gpt-4o-mini & 0.46 \space\space ± 0.02 & 0.56 \space\space ± 0.02 & 0.64 \space\space ± 0.01 & 0.40 \space\space ± 0.03 \\
llama4-maverick & 0.65 \space\space ± 0.01 & 0.60 \space\space ± 0.01 & 0.57 \space\space ± 0.00 & 0.53 \space\space  ± 0.03 \\
o1 & 0.69 \space\space ± 0.01 & 0.70 \space\space ± 0.02 & 0.83 \space\space ± 0.02 & 0.83 \space\space ± 0.03 \\
o3 & 0.74 \space\space ± 0.02 & 0.76 \space\space ± 0.02 & 0.78 \space\space ± 0.01 & 0.79 \space\space ± 0.02 \\
o4-mini & 0.72 \space\space ± 0.02 & 0.73 \space\space ± 0.02 & 0.80 \space\space ± 0.02 & 0.78 \space\space ± 0.02 \\
HallOumi-8B & -- & 0.78 \space\space  ± 0.00 & -- & -- \\
HallOumi-8B-Clf & 0.74 \space\space ± 0.00 & -- & -- & -- \\
\bottomrule
\end{tabular}
\caption{Mean and standard deviation of recall scores across 17 models under 4 prompting modes. Boldfaced scores indicate the best mean within each prompt mode.}
\label{tab:recall}
\end{table}

\end{document}